\PassOptionsToPackage{table}{xcolor}
\documentclass[runningheads]{llncs}

\usepackage{eccv}

\makeatletter

\newcommand{\appendixtitle}[1][\@title]{%
  \par
  \newpage
  \null
  \begin{center}%
    {\Large\bfseries\boldmath
     \pretolerance=10000
     #1\\[.3em]{\large\bfseries Supplementary Material}}%
  \end{center}%
  \vskip .8cm\par%
}
\makeatother

\usepackage{eccvabbrv}

\usepackage{wrapfig}
\usepackage{graphicx}
\usepackage{booktabs}
\usepackage{algorithm}
\usepackage{algpseudocode}

\usepackage[accsupp]{axessibility}  

\usepackage{hyperref}

\definecolor{best}{RGB}{244,207,199}
\definecolor{secbest}{RGB}{255,232,210}
\definecolor{cvprblue}{rgb}{0.21,0.49,0.74}
\newcommand{\ours}{NoPA\xspace}

\definecolor{red_sj}{RGB}{223,105,80}
\definecolor{blue_sj}{RGB}{92,102,240}
\definecolor{green_sj}{RGB}{47,170,148}

\usepackage{multirow} 

\usepackage{orcidlink}

\begin{document}

\title{NoPA: Non-Parametric Online 3D Scene Graph Generation}

\author{Qi Xun Yeo\inst{1} \and
Seungjun Lee\inst{1} \and
Yan Li\inst{1} \and
Gim Hee Lee\inst{1}}

\authorrunning{Yeo et al.}

\institute{Department of Computer Science, National University of Singapore\\
\email{\{qixunyeo, seungjun.lee\}@u.nus.edu}, 
\email{\{yan.li,gimhee.lee\}@nus.edu.sg}}

\maketitle

\begin{figure}[t]
  \centering
  \includegraphics[width=\textwidth]{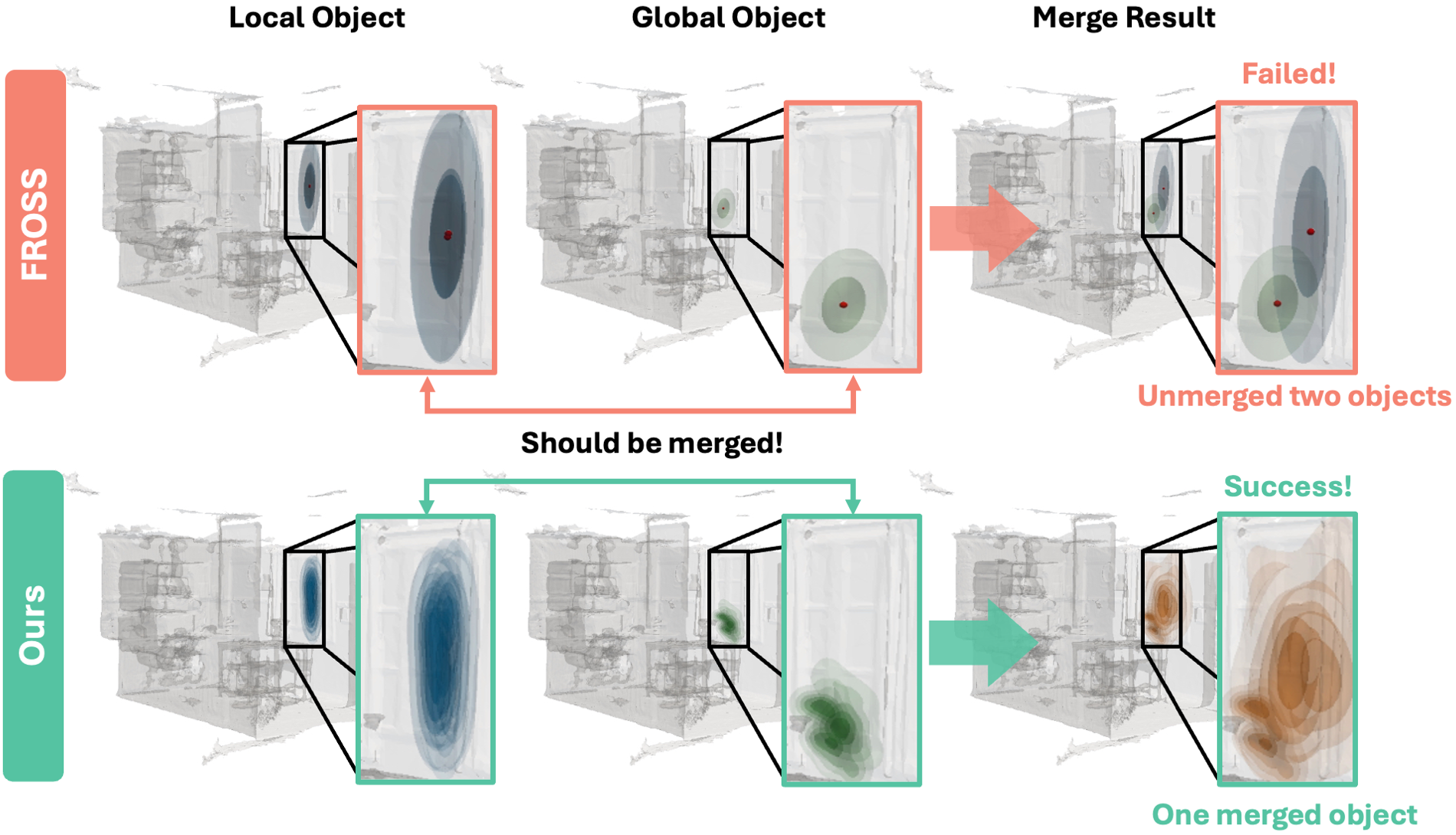}
  \vspace{-15pt}
  \caption{
  \textbf{Gaussian (FROSS~\cite{hou2025fross}) vs Non-Parametric (Ours).} We show the problem of under-merging when merging objects represented as 3D Gaussians on a single window object instance.
  \textbf{Top Row (\textcolor{red_sj}{FROSS}):} Under Gaussian parameterization, the local object (1st column) fails to merge with the global object (2nd column) which results in under-merging (3rd column). After the merging process, the window instance is incorrectly represented by two smaller Gaussians instead of a single larger one. This fragmentation produces splintered 3D scene graphs that do not reflect the true scene structure. Strict post-processing filters would then remove fragmented objects together with their relations. As a result, the quality of the final 3D SSG degrades.
  \textbf{Bottom Row (\textcolor{green_sj}{Ours}):} Our non-parametric formulation allows the local object to merge with the global object and preserves richer geometric support. This reduces under-merging and produces more accurate and consistent 3D SSGs.
  }
  \vspace{-20pt}
  \label{fig:teaser}
\end{figure}

\vspace{-20pt}
\begin{abstract}
Classic 3D scene graph generation approaches fail to work in real-time due to the heavy computational cost of environment mapping and the need to generate intermediate point-cloud representations.
To alleviate this issue, a recent work eschews point clouds in favor of a lightweight Gaussian distribution for each object. This approximation drastically speeds up inference and enables real-time 3D scene graph generation. However, the representation has two key weaknesses. \textbf{1)} Each object is approximated by a single 3D Gaussian, which causes a severe loss of 3D geometric detail. \textbf{2)} The discrepancy between this approximation and the true object geometry exacerbates the inaccurate merging of object candidates during online inference.
To address these issues, we propose \textbf{NoPA}, which represents each object as a separate non-parametric distribution. This formulation retains 3D geometric information while preserving real-time inference of the parametric Gaussian formulation. To build upon our novel object representation, we propose a tailored merging strategy to recover coherent object instances. Specifically, we leverage maximum mean discrepancy on kernel density estimates to enable robust merging of object candidates during online exploration while minimizing added computational complexity. The key is to maintain a fixed particle set per object.
Furthermore, to rectify the relation loss caused by misclassified objects, NoPA propagates relationships between objects with high affinity. Experiments show that NoPA substantially outperforms current methods without sacrificing real-time inference speed.

\end{abstract}

\section{Introduction}\label{sec:intro}
We study \emph{online 3D scene graph generation} from 
streaming RGB-D images. A 3D semantic scene graph (3D SSG) encodes objects and their relationships in a structured representation and serves as a critical abstraction for embodied AI and robotics. It enables downstream tasks such as navigation~\cite{10.1109/HUMANOIDS47582.2021.9555790, DBLP:journals/ral/ChangBC23, yin2024sgnav, yin2025gcvln, werby23hovsg, huang2026msgnavunleashingpowermultimodal, wang2025dynam3d,wang2025d3d}, scene generation~\cite{graph2scene2021, zhai2023commonscenes, zhai2024echo, yang2025mmgdreamer, Liu_2025_ICCV}, and manipulation~\cite{graph2scene2021, gu2024conceptgraphs, yin2025gcvln, buechner2026momasg, lee2024segment}. These capabilities are essential in medical~\cite{ege2022, DBLP:conf/miccai/OzsoyCHPN23, guo2024tri}, construction~\cite{DBLP:conf/eccv/CelenHSGAOW24, lin2024instructscene, Nyffeler_2025_ICCV,lee2025diet}, and autonomous driving domains~\cite{greve2023curb, fischer2024multi, ijcai2025p270, lee2026segment}.

Despite its importance, most prior works address 3D SSG generation in an offline setting without strict real-time constraints~\cite{armeni20193d, wald2020learning, wald2022learning, zhang2021exploiting, wang2023vl, wu2021scenegraphfusion, feng20233d, Yeo_2025_ICCV}. To the best of our knowledge, only three works tackle the online setting~\cite{wu2021scenegraphfusion, wu2023incremental, hou2025fross} with real-time performance.
SceneGraphFusion~\cite{wu2021scenegraphfusion} and MonoSSG~\cite{wu2023incremental} rely on simultaneous localization and mapping (SLAM) pipelines to reconstruct geometry before scene graph prediction, which introduces significant computational overhead and limits scalability.
FROSS~\cite{hou2025fross} avoids explicit mapping by lifting 2D scene graphs into 3D and achieves high frame rates by approximating each object as a Gaussian distribution.

Although FROSS achieved real-time performance without SLAM, their Gaussian parameterization imposes a restrictive geometric assumption where each object is modeled as an ellipsoid defined by its covariance. This approximation discards fine geometric structure and makes merging fragile. As shown in \cref{fig:teaser}, thin or planar structures such as pictures and windows often produce near-singular covariance matrices leading to under-merging. Different viewpoints of the same object often yields Gaussian ellipsoids with inconsistent covariances and spatial offsets causing incorrect merges. Consequently, merging decisions are prone to be unstable since incorrect merges and undermerges accumulate over time and progressively degrade the global 3D SSG. These limitations arose fundamentally from the parametric assumption instead of implementation details.

To overcome both the computational burden of SLAM-based pipelines and the geometric limitations of Gaussian modeling, we introduce \textbf{NoPA} (\textbf{Non}-\textbf{PArametric} Online 3D Scene Graph Generation). Our NoPA replaces Gaussian object modeling with a fixed-size non-parametric particle set that preserves geometric support. This formulation removes the restrictive ellipsoidal assumption while maintaining constant memory and runtime complexity. Merging two object candidates proceeds by estimating a kernel density over their unified particle support and resampling a fixed-size set from it. This integrates multi-view geometric evidence while preserving a constant-size representation. Consequently, NoPA matches the real-time efficiency of SLAM-free parametric methods while retaining substantially richer geometric structure.

Adopting a non-parametric representation shifts the merging problem from covariance comparison to distribution comparison across views. We empirically find that covariance similarity is insufficient and unreliable under viewpoint variation, often resulting in under-merging (\cf \cref{fig:teaser}). We address this by introducing a principled distribution-level merging criterion based on Maximum Mean Discrepancy (MMD). MMD measures similarity directly between particle sets in feature space and provides a stable signal even when geometric support differs across views or when 2D predictions are noisy. This significantly improves merging robustness in ambiguous cases and prevents cascading structural errors in the final 3D SSG. To further strengthen the global consistency of the 3D SSG, our NoPA incorporates a relationship propagation mechanism as a post-processing step. We cluster object candidates using the previously computed MMD scores and propagate relationships across clusters to recover missing edges. This mitigates structural damage caused by imperfect merging and enhances overall graph completeness without sacrificing runtime efficiency.

In summary, our main contributions are as follows:
\begin{itemize}
    \item We introduce \textbf{NoPA}, a non-parametric formulation for online 3D scene graph generation that eliminates restrictive Gaussian assumptions while preserving fixed memory usage and real-time computational complexity.
    \item We design a principled \textbf{distribution-level merging framework} based on Maximum Mean Discrepancy that replaces fragile covariance similarity and improves robustness under viewpoint variation and noisy predictions.
    \item We develop a \textbf{relationship propagation mechanism} guided by distribution similarity to recover missing relations and reinforce graph consistency.
    \item We achieve state-of-the-art performance on multiple online 3D SSG benchmarks while maintaining competitive real-time efficiency.
\end{itemize}

\section{Related Work}

\noindent \textbf{Offline 3D SSG.} 
Offline 3D SSG generation approaches aim to estimate a 3D scene graph using ground truth 3D geometry~\cite{armeni20193d, wald2020learning, wald2022learning, zhang2021exploiting, wang2023vl} or multi-view RGB-D images~\cite{DBLP:journals/access/SonogashiraIK22, feng20233d, gu2024conceptgraphs, zhang2025fungraph3d, Yeo_2025_ICCV} in a non-incremental manner. Wald et al. \cite{wald2020learning} first proposed the problem of 3D SSG generation and attempted to solve it by modeling pairwise relationships to predict the graph. Most modern approaches rely on multi-view RGB-D images. Wang et al. \cite{wang2023vl} leverage pretrained model priors by distilling knowledge from a multimodal oracle model into a 3D model. Yeo et al. \cite{Yeo_2025_ICCV} propose a statistical confidence rescoring mechanism to refine low-confidence predictions and use SegmentAnything (SAM)~\cite{kirillov2023segment} instance masks to enhance node features. Koch et al.~\cite{koch2024open3dsg} distill knowledge from visual language models (VLMs) into a 3D graph neural network (GNN). Gu et al. \cite{gu2024conceptgraphs} run a class-agnostic segmentation model to obtain candidate objects, associate them across views using geometric and semantic similarity, instantiate nodes in a 3D SSG refined by VLMs, and prompt a large language model (LLM) with object pairs to infer spatial relations. Koch et al.~\cite{koch2025relationfield} build a relationship-aware 3D representation that supports node and predicate queries. Zhang et al.~\cite{zhang2025fungraph3d} introduce functional relationships and contribute a functional 3D SSG dataset.
These offline systems typically aggregate information over a fixed set of frames and perform expensive global association and refinement. This becomes challenging under strict online latency and bounded-memory constraints. 
In contrast, our NoPA targets online 3D SSG generation with constant memory per object. We replace Gaussian object modeling with fixed-size particle sets, use a distribution-level merging criterion to improve cross-view association, and propagate relations within affinity clusters to recover missed edges during incremental fusion.

\smallskip
\noindent \textbf{Online 3D SSG.} The first work to tackle 3D SSG generation in an online setting is by Kim et al.~\cite{kim2019graph3d}. The work focuses on predicting local 3D SSGs that combine into a global 3D SSG. However, it fails to reach real-time speeds required for practical deployment. Wu et al.~\cite{wu2021scenegraphfusion} introduce a graph convolutional network (GCN) based aggregation function, abbreviated as FAN, to improve the predicted 3D SSG while running RGB-D SLAM to obtain dense intermediate 3D representations. MonoSSG~\cite{wu2023incremental} proposes an entity association approach that lifts 2D entities into 3D, enhances ORBSLAM, and introduces a geometric gate that fuses geometric information with multi-view image features. More recently, FROSS~\cite{hou2025fross} approximates objects as 3D Gaussians to avoid heavy point cloud processing or environment mapping, which accelerates inference. However, by removing precise localization, it fails to use geometric information available in 3D for merging and instead relies on an approximation of the 2D object shape from RGB-D observations. In contrast, our NoPA balances the trade-off between preserving geometric detail and improving inference speed through a fixed number of particles per object. This design maintains real-time performance while retaining richer 3D geometry than FROSS, which improves merging and overall model performance.

\section{Problem Definition}
Given $N$ multi-view RGB images of a 3D scene, denoted as $\{I_i\}_{i=1}^{N}$, our goal is to estimate a 3D scene graph:
\begin{equation}
    G^{3D} = (O, R),
\end{equation}
where $O=\{o_j\}_{j=1}^{M}$ is the set of $M$ object nodes and $R=\{r_{k\rightarrow j}\}_{j,k=1}^{M}$ is the set of directed relationship edges over object pairs. 
Node $j$ has an object (category) label $o_j$, and the directed edge from node $k$ to node $j$ has a predicate label $r_{k\rightarrow j}$. 
Equivalently, the scene graph can be represented as a set of triplets $\{(o_k, r_{k\rightarrow j}, o_j)\}$.

We study the \emph{online} setting. The method does not assume access to the full image set $\{I_i\}_{i=1}^{N}$ at test time. Instead, it receives a sequential stream of partial observations and incrementally updates $G^{3D}$ as new images arrive during scene exploration.

\section{Preliminaries}
\label{sec:preliminaries}
We build our framework based on FROSS~\cite{hou2025fross}, where it updates the scene graph incrementally from a stream of RGB frames and avoids explicit environment mapping.
Given RGB observations $\{I_i\}_{i=1}^{N}$, FROSS first predicts a per-frame 2D scene graph
$G^{2D}_i = g_{\phi}(I_i)$, where $g_{\phi}$ is a pretrained 2D SSG detector.
It then lifts this 2D graph into the world frame using the per-frame depth map $d_i$
and camera pose $P_i\in SE(3)$, and fuses the lifted result into the global 3D scene graph:
\begin{equation}
    G^{3D}_i = \mathcal{B}\!\left(G^{2D}_{i}\mid d_i, P_i\right)\odot G^{3D}_{i-1},
    \label{eq:updateEq}
\end{equation}
where $G^{3D}_{i-1}$ and $G^{3D}_i$ denote the global 3D scene graph before and after processing frame $i$.
The operator $\mathcal{B}(\cdot\mid d_i,P_i)$ maps each 2D node in $G^{2D}_i$ to a 3D object hypothesis by back-projecting image evidence with $d_i$ and transforming it to the world frame with $P_i$.
The fusion operator $\odot$ performs data association between the lifted hypotheses and existing nodes in $G^{3D}_{i-1}$, followed by merging and state updates.

FROSS represents each 3D object node $o_j$ with a single Gaussian in $\mathbb{R}^3$:
\begin{equation}
    p(\mathbf{x}\mid o_j)=\mathcal{N}\!\left(\mathbf{x};\mu_j,\Sigma_j\right),
\end{equation}
where $\mathbf{x}\in\mathbb{R}^3$ is a 3D point, $\mu_j\in\mathbb{R}^3$ is the object centroid, and $\Sigma_j\in\mathbb{R}^{3\times 3}$ is the covariance.
This Gaussian is initialized by lifting a 2D Gaussian estimated from the predicted 2D bounding box.
During fusion, it merges a lifted object hypothesis $i$ with an existing global object $j$
when their semantic labels match and their Hellinger distance $d_H(i,j)$ falls below a threshold $\delta_H$.
Approximating each object with a Gaussian $\mathcal{N}(\mu,\Sigma)$, the Hellinger distance admits a closed form:
\begin{equation}
d_H(i,j)=\sqrt{1-\exp\!\big(-d_B(i,j)\big)}.
\end{equation}
where $d_B(i,j)$ is the Bhattacharyya distance between two Gaussians
$\mathcal{N}(\mu_i,\Sigma_i)$ and $\mathcal{N}(\mu_j,\Sigma_j)$:
\begin{equation}
\begin{split}
d_B(i,j)
&=\frac{1}{8}\Delta\mu_{ij}^{\top}\Sigma^{-1}\Delta\mu_{ij}
+\frac{1}{2}\ln\!\Bigg(\frac{\det\Sigma}{\sqrt{\det\Sigma_i\,\det\Sigma_j}}\Bigg),\\
\Delta\mu_{ij}&=\mu_i-\mu_j,\qquad
\Sigma=\frac{\Sigma_i+\Sigma_j}{2}.
\end{split}
\end{equation}

\medskip
\noindent \textbf{Limitations of FROSS.} FROSS represents each object with a single Gaussian, which coarsely approximates the object geometry with an ellipsoidal shape, removing the fine structural details of the instances. This approximation loss is accumulated across the streaming images, resulting in disjoint object instances from fragile merging and incorrect associations. 
\vspace{-7pt}
\section{Our Method}
\label{sec:method}
\vspace{-5pt}

\begin{figure}[t]
  \centering
  \includegraphics[width=\textwidth]{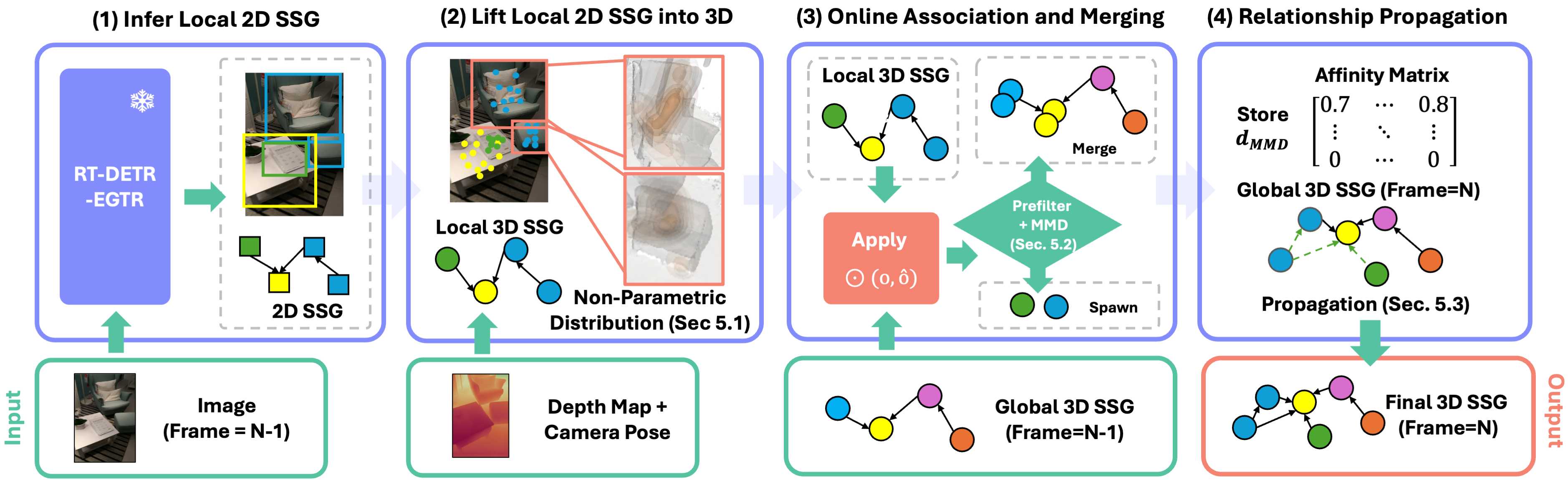}
  \caption{
  Overview of our online 3D scene graph generation pipeline.
  (1) A pretrained RT-DETR-EGTR \cite{Im_2024_CVPR, Zhao_2024_CVPR} model predicts a local 2D scene graph from each RGB frame.
  (2) For every object node, we sample pixels inside its 2D bounding box, back-project them with depth, and obtain a 3D particle set in the world frame.
  (3) We associate each local particle set with existing global objects using a two-stage test: a constant-time Hellinger pre-filter followed by MMD on ambiguous cases.
  (4) We propagate relations within high-affinity clusters to reduce relation dropouts from the 2D predictor.
  }
  \label{fig:architecture}
  \vspace{-15pt}
\end{figure}

\noindent \textbf{Overview.}
\cref{fig:architecture} summarizes our online 3D scene graph generation framework.
We improve FROSS by replacing its single-Gaussian object model with a non-parametric particle representation, and redesigning online fusion around distributional similarity.
At timestep $i$, a pretrained RT-DETR-EGTR model infers a \emph{local 2D scene graph} $G^{2D}_i$ from the RGB frame $I_i$.
$G^{2D}_i$ contains object nodes (2D boxes with class labels) and relation edges (pairwise predicates) in the image plane.
We lift each detected 2D object node into a 3D particle set using the depth map $d_i$ and camera pose $P_i$ (\cref{nonpara}), and fuse these local 3D candidates into the global 3D scene graph from the previous timestep (\cref{merging}).
Fusion uses a fast two-stage association rule: a constant-time Hellinger pre-filter for clear matches/mismatches, and a maximum mean discrepancy (MMD) test for borderline pairs.
We then stabilize the relation set by propagating relations within affinity clusters, which helps to recover the relations missed by the 2D predictor in a single view (\cref{prop}).
This design preserves object support in $\mathbb{R}^3$, improves cross-view association, and maintains constant memory per object.

\vspace{-3pt}
\subsection{Non-Parametric Object Representation}
\label{nonpara}

\noindent \textbf{From 2D boxes to 3D particles.}
For each detected 2D object node in $G^{2D}_i$ with bounding box $b$, we sample $n$ pixels $\{\mathbf{u}_k\}_{k=1}^{n}$ uniformly within $b$.
Using the depth map $d_i$, we back-project each pixel to a camera-frame 3D point
$\mathbf{X}^{c}_k=\pi^{-1}(\mathbf{u}_k,d_i(\mathbf{u}_k))$ and transform it to the world frame with the camera pose $P_i\in SE(3)$:
\begin{equation}
\mathbf{x}_k = P_i \, \mathbf{X}^{c}_k,\qquad \mathbf{x}_k\in\mathbb{R}^3 .
\end{equation}
The lifted object is represented by the particle set $\mathcal{X}(o)=\{\mathbf{x}_k\}_{k=1}^{n}$.

\medskip
\noindent \textbf{Kernel density view.}
We treat the particle set as samples from an unknown object occupancy distribution and form a kernel density estimate (KDE):
\begin{equation}
\hat{f}(\mathbf{x}\mid o)=\frac{1}{n}\sum_{k=1}^{n}\kappa(\mathbf{x},\mathbf{x}_k),
\end{equation}
where $\kappa(\cdot,\cdot)$ is an RBF kernel:
\begin{equation}
\kappa(\mathbf{x},\mathbf{y})=\exp\!\Big(-\tfrac{\|\mathbf{x}-\mathbf{y}\|_2^2}{2\sigma^2}\Big).
\end{equation}

\begin{figure}[t]
  \centering
  \includegraphics[width=\textwidth]{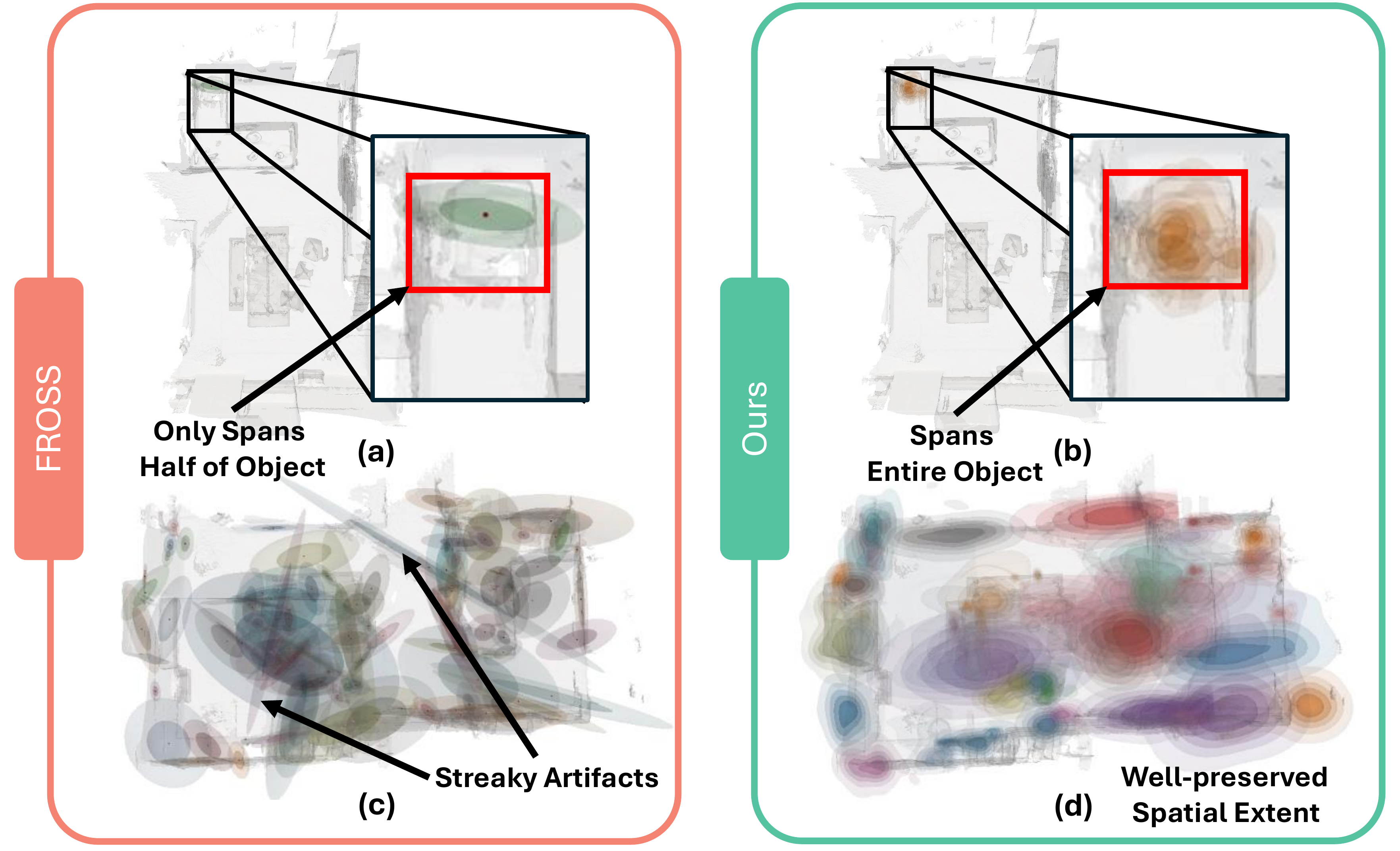}
  \caption{The visualization of objects in Scene \texttt{41385849} from the 3DSSG dataset. In the top row, we visualize an instance of the sink class localized by a red bounding box. In the bottom row, we visualize the global 3D object instances. \textbf{Left:} Visualized Gaussian blobs from FROSS\cite{hou2025fross}. The Gaussian blob only encompasses half of the sink in (a). Spurious blobs spanning across the scene in (c) visualizes the impact of incorrect merging. \textbf{Right:} Visualized kernel densities for our particle set. The entire sink is associated with our particle set in (b). Our representation attains good coverage on objects across the scene without large artifacts in (d).}
  \label{fig:filter_viz}
\end{figure}

\medskip
\noindent \textbf{Remark.}
A single 3D Gaussian enforces an ellipsoidal prior, which blurs multi-part structures and amplifies approximation error across views.
As shown in Fig.~\ref{fig:filter_viz}, a particle set preserves object support in $\mathbb{R}^3$ and admits multi-modality, which reduces over-merging and under-merging under partial observations.

\subsection{Online Association and Merging}
\label{merging}

We maintain a set of global objects, each with particles $\mathcal{X}(o)$.
Given a local object candidate $\hat{o}$ with particles $\mathcal{X}(\hat{o})$ and an existing global object $o$ with particles $\mathcal{X}(o)$, we decide whether $\hat{o}$ corresponds to $o$ and should be merged, or whether it should spawn a new global object.
Our association uses a two-stage criterion: 1) a \textit{\textbf{constant-time Hellinger pre-filter}} that resolves clear cases, followed by 2) an \textit{\textbf{MMD test}} on ambiguous pairs.

\medskip
\noindent \textbf{Stage 1: Constant-time pre-filter.}
To obtain a cheap moment-level proxy, we fit a \emph{unimodal Gaussian} to each particle set by matching its first two moments:
$(\mu,\Sigma)$ for $o$ and $(\hat{\mu},\hat{\Sigma})$ for $\hat{o}$.
We then compute the Hellinger distance $d_H$ between the two fitted Gaussians.
Instead of committing to a hard threshold at $\delta_H$, we introduce a \emph{margin band} of width $2\epsilon$:
\begin{equation}
\text{merge if } d_H < \delta_H-\epsilon,\qquad
\text{spawn if } d_H > \delta_H+\epsilon.
\end{equation}
The band prevents unstable decisions when $d_H$ fluctuates under depth noise, truncation, or limited view overlap.
Pairs within $[\,\delta_H-\epsilon,\ \delta_H+\epsilon\,]$ remain undecided and move to Stage~2.

\medskip
\noindent \textbf{Stage 2: MMD for ambiguous pairs.}
For candidates inside the margin band $[\,\delta_H-\epsilon,\ \delta_H+\epsilon\,]$, we compute the maximum mean discrepancy (MMD) between the two KDEs:
\begin{align}
d_{\mathrm{MMD}}^2(o,\hat{o})
&=
\mathbb{E}_{\mathbf{x},\mathbf{x}'\sim \mathcal{X}(o)}
\!\left[\kappa(\mathbf{x},\mathbf{x}')\right]
+\mathbb{E}_{\mathbf{y},\mathbf{y}'\sim \mathcal{X}(\hat{o})}
\!\left[\kappa(\mathbf{y},\mathbf{y}')\right] \nonumber\\
&\quad
-2\,\mathbb{E}_{\mathbf{x}\sim \mathcal{X}(o),\,\mathbf{y}\sim \mathcal{X}(\hat{o})}
\!\left[\kappa(\mathbf{x},\mathbf{y})\right].
\end{align}
We set $\sigma^2$ with the median heuristic over random pairs from $\mathcal{X}(o)\cup\mathcal{X}(\hat{o})$.
We merge if $d_{\mathrm{MMD}}(o,\hat{o})\le \delta_{\mathrm{MMD}}$, and spawn otherwise, where $\delta_{\mathrm{MMD}}$ is a fixed threshold calibrated on a held-out sequence to match the desired precision--recall trade-off.

\medskip
\noindent \textbf{Why MMD?}
The Stage~1 Gaussian fit is intentionally coarse, where different particle sets can share similar $(\mu,\Sigma)$ under partial views, thin structures, or multi-part objects.
MMD directly compares the \emph{full distributions} induced by the KDEs in a reproducing kernel Hilbert space, which makes it sensitive to support mismatch beyond first- and second-order moments.
It is also model-free and works naturally with our particle representation, with the requirement of only kernel evaluations without meshing or explicit points correspondence.

\medskip
\noindent \textbf{Decision rule.}
Following the online update in Eq.~\ref{eq:updateEq}, the fusion operator $\odot$ implements the two-stage association between a local candidate $\hat{o}$ and a global object $o$:
\begin{equation}
\odot(o,\hat{o})=
\begin{cases}
\text{merge} & d_H < \delta_H-\epsilon,\\
\text{spawn} & d_H > \delta_H+\epsilon,\\
\text{merge} & \text{otherwise and } d_{\mathrm{MMD}}(o,\hat{o})\le \delta_{\mathrm{MMD}},\\
\text{spawn} & \text{otherwise.}
\end{cases}
\end{equation}
It first accepts or rejects unambiguous pairs using $d_H$, and invokes $d_{\mathrm{MMD}}$ only within the margin band to resolve borderline cases where moment matching proves unreliable.

\medskip
\noindent \textbf{Merge update with constant memory.}
After each merge, we take the union of the particles, fit a KDE over the union support, and resample a fixed-size set of $n$ particles:
\begin{equation}
\mathcal{X}(o) \leftarrow \mathrm{Resample}_n\!\big(\mathcal{X}(o)\cup \mathcal{X}(\hat{o})\big).
\label{eq:resample}
\end{equation}
This step preserves geometric information from both candidates and prevents particle growth over time.

\begin{figure}[t]
  \centering
  \includegraphics[width=\textwidth]{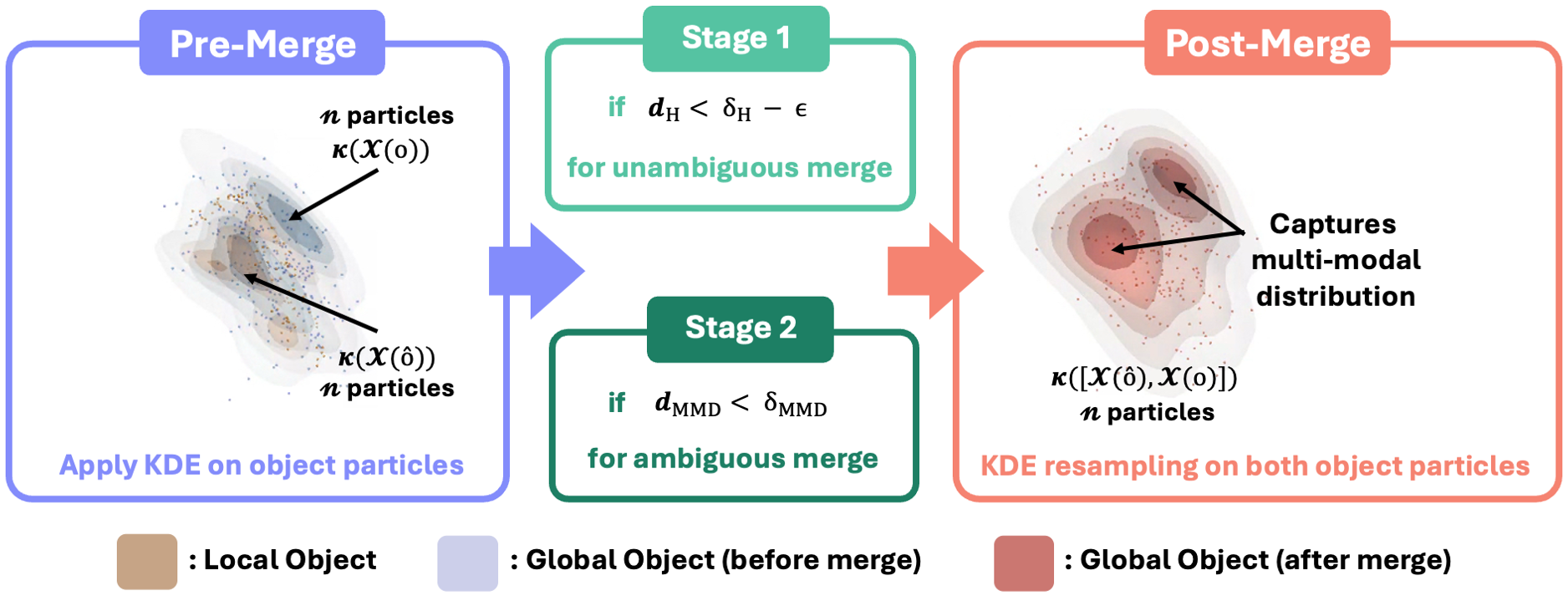}
  \caption{
  Visualization of the merging process for an object in Scene \texttt{7272e16c} from the 3DSSG dataset. 
  If the local particle set and the global particle set yields a small Hellinger distance (\(d_H < \delta_H - \epsilon\)) after fitting a unimodal Gaussian (\textit{Stage 1}), their covariances clearly matches, and the merge decision is straightforward. If the Hellinger distance falls within the margin band (\textit{Stage 2}), covariance alignment alone is insufficient to determine merging. Consequently, we apply the more robust MMD criterion and merge the particle sets if 
  \(d_{\mathrm{MMD}} < \delta_{\mathrm{MMD}}\).
  \textcolor{blue_sj}{\textbf{Left:}} Initial kernel density estimates of the local and global particle sets. \textcolor{red_sj}{\textbf{Right:}} Kernel density estimate of the merged particle set after fusion. The number of particles remain constant after merging due to KDE resampling while the updated particle set captures the multi-modal distribution.}
  \label{fig:merge_viz}
\end{figure}

\medskip
\noindent \textbf{Remark.}
The Hellinger pre-filter keeps most association decisions inexpensive.
MMD focuses computation on hard cases where first- and second-order moments agree but distribution support differs.
As shown in Fig.~\ref{fig:merge_viz}, resampling maintains constant memory and avoids collapsing the representation toward a single mode, which stabilizes long-horizon fusion.

\subsection{Relationship Propagation with Affinity Clusters}
\label{prop}

Online 3D SSG generation is sensitive to relation dropouts, where low-confidence predicate edges are suppressed, and a missing edge in one frame may never be recovered later.
To increase robustness, we exploit geometric redundancy across objects with similar 3D support.
Specifically, we reuse the MMD scores computed during ambiguous association to build an affinity matrix over object candidates.
Low MMD indicates high geometric similarity, which defines affinity clusters.
For a newly merged or spawned node, we first copy its observed 2D relations to the corresponding 3D node.
We then propagate candidate relations within its affinity cluster and finalize the relation type by majority voting on the available evidence.
This aggregation recovers relations missed in a single view, reduces sensitivity to relation confidence thresholds, and limits relation drift.

\medskip
\noindent \textbf{Remark.}
Affinity-based propagation recovers relations that are missed in a single view by borrowing consistent evidence from geometrically similar neighbors.
Majority voting reduces the impact of occasional false positives and prevents relation drift.

\section{Experiments}
\label{sec:Experiments}
\subsection{Experimental Setup}
\noindent \textbf{Datasets.} 
We evaluate on the 3DSSG~\cite{wald2020learning} and ReplicaSSG datasets following the same evaluation set up as FROSS~\cite{hou2025fross}. 3DSSG contains 1482 scenes which are annotated with 21974 objects and 16324 predicate relationship between two objects. Each scene encompasses multiple short videos with different motion trajectories capturing the entire scene. The dataset is challenging due to the poor quality of images captured with significant motion blur. ReplicaSSG contains 18 scenes which are annotated with 1526 objects and 582 predicate relationship between two objects. The dataset contains high quality images and meshes inherited from the Replica dataset. The large number of objects and predicates per scene makes evaluation non-trivial.

\medskip
\noindent \textbf{Baseline Methods.}
For 3DSSG, we compare our method with the other online 3D SSG generation methods that utilize RGB-D images with the ground truth pose information including Kim's framework\cite{kim2019graph3d}, JointSSG\cite{wu2023incremental}, FROSS\cite{hou2025fross}. We exclude MonoSSG\cite{wu2023incremental} since it only uses RGB images. We reproduce the approaches using their respective GitHub repositories\footnote{For JointSSG, we follow https://github.com/ShunChengWu/3DSSG. For Kim's framework and FROSS, we follow https://github.com/Howardkhh/FROSS.}.

For ReplicaSSG, we only compare with FROSS since it is the only prior approach that evaluates on the dataset.

\medskip
\noindent \textbf{Implementation Details.}
We choose to keep the top 20 relationships from the pretrained RT-DETR-EGTR instead of the top 10 to reduce the loss of the relationship edge from the final 3D SSG when running FROSS. This ensures a fair comparison between the competing approaches.

Our implementation is built using the PyTorch framework. Following FROSS \cite{hou2025fross}, we pretrain the initial 2D SSG generation model - RT-DETR-EGTR for a maximum of 50 epochs. 
All experiments are conducted on a single RTX 3090 GPU for a fair comparison. We utilized \(n=256\) particles to represent each object. \(\delta_{\mathrm{MMD}}\) is set to 0.7 for 3DSSG and 0.6 for ReplicaSSG. \(\epsilon\) is kept at 0.05. 

Similar to FROSS, we omit the \(None\) class for both object and predicate predictions that were previously implemented for SceneGraphFusion~\cite{wu2021scenegraphfusion}. The advantage of removing the \(None\) class is the prevention of possible overfitting to the \(None\) class which is the most prevalent ground truth annotation. 

We follow the same strict criteria as FROSS to match the predicted object candidates to the ground truth object instances: (1) The majority of points (more than 50\%) sampled from our continuous particle set should have its nearest ground truth point mapped to the corresponding matched ground truth object. (2) The fraction of overlap counts belonging to the second-largest matched ground truth object compared to the overlap counts belonging to the largest matched ground truth object must not exceed 75\%. These criteria enforces one-to-one correspondence between predicted and ground truth objects which increases the robustness of the evaluation.

\medskip
\noindent \textbf{Evaluation Metrics.} 
In terms of evaluating the 3D scene graph centric performance, we follow SceneGraphFusion\cite{wu2021scenegraphfusion} and MonoSSG\cite{wu2023incremental} to report the overall \textbf{top-1} recall (Recall) for the object class estimation (Obj.), the predicate estimation (Pred.), and the relationship triplet estimation (Rel.). We also report the mean recall (mRecall) for the object class estimation (Obj.), the predicate estimation (Pred.) only. In terms of evaluating the runtime efficiency for the online setting, we report the latency as in \cite{hou2025fross} and additionally compare the memory requirements of each method.

\subsection{Quantitative Results}

\begin{table}[t]
    \centering
    \caption{Comparison with state-of-the-art online 3D SSG generation approaches on the 3DSSG dataset with 20 object classes and 7 predicate classes. The top group of results are the reported results from the respective papers. The middle group of results marked with the \(\dagger\) are the reproduced results from the respective GitHub repositories. The \colorbox{best}{\textbf{Best}} and \colorbox{secbest}{Second Best} results are highlighted, respectively.}
    
    \renewcommand{\arraystretch}{0.92} 
    \setlength{\abovecaptionskip}{4pt}
    \setlength{\belowcaptionskip}{-4pt}
    
    \begin{tabular*}{\linewidth}{@{\extracolsep{\fill}}c|ccc|cc|c|c}
    \toprule
      \multirow{2}{*}{Method} & \multicolumn{3}{c|}{Recall\% (\(\uparrow\))} & \multicolumn{2}{c|}{mRecall\% (\(\uparrow\))} & \multirow{2}{*}{Latency (\(ms\downarrow\))} & \multirow{2}{*}{VRAM (MB\(\downarrow\))} \\
      \cline{2-6}
       & Rel. & Obj. & Pred. & Obj. & Pred. &  &  \\
      \hline
       JointSSG \cite{wu2023incremental} & 25.5 & 58.1 & 27.3 & 43.0 & 33.3 & 191 & - \\
       Kim \cite{kim2019graph3d} & 9.1 & 59.0 & 7.1 & 51.0 & 8.0 & 310 & - \\
       FROSS \cite{hou2025fross} & 27.9 & 62.4 & 33.0 & 63.8 & 18.0 & 7 & - \\
       \hline
       JointSSG\(\dagger\) \cite{wu2023incremental} & 23.4 & 55.4 & 27.0 & 45.4 & \cellcolor{best}\textbf{35.3} & 284 & 3252 \\
       Kim\(\dagger\)\footnotemark \cite{hou2025fross} & 0.9 & 52.1 & 1.1 & 44.2 & 0.4 & 488 & \cellcolor{best} \textbf{1204} \\
       FROSS\(\dagger\) \cite{hou2025fross} & 25.7 & 60.6 & 30.7 & 62.4 & 17.7 & \cellcolor{best}\textbf{22} & \cellcolor{best} \textbf{1204} \\
       \hline
       Ours(\(n=128\)) & \cellcolor{secbest}49.9 & \cellcolor{secbest}68.5 & \cellcolor{secbest}58.5 & \cellcolor{secbest}65.7 & \cellcolor{secbest}30.2 & \cellcolor{secbest}26 & \cellcolor{secbest} 1206 \\
       Ours(\(n=256\)) & \cellcolor{best}\textbf{53.2} & \cellcolor{best}\textbf{69.0} & \cellcolor{best}\textbf{61.4} & \cellcolor{best}\textbf{66.4} & 29.4 & 27 & \cellcolor{secbest} 1206 \\
    \bottomrule
    \end{tabular*}
    
    \label{tab:main}
\end{table}

\footnotetext{Kim's method crashed in the 132nd scene out of 157 scenes from RAM OOM. The point clouds stored are further resized down by 5x to fit the memory possibly leading to the large drop in performance.}

\begin{table}[t]
    \centering
    \caption{Comparison with FROSS on the ReplicaSSG dataset with 34 object classes and 9 predicate classes. \(\dagger\) refers to the reproduced results. The \colorbox{best}{\textbf{Best}} and \colorbox{secbest}{Second Best} results are highlighted, respectively.}
    
    \renewcommand{\arraystretch}{0.92} 
    \setlength{\abovecaptionskip}{4pt}
    \setlength{\belowcaptionskip}{-4pt}
    
    \begin{tabular*}{\linewidth}{@{\extracolsep{\fill}}c|ccc|cc|c|c}
    \toprule
      \multirow{2}{*}{Method} & \multicolumn{3}{c|}{Recall\% (\(\uparrow\))} & \multicolumn{2}{c|}{mRecall\% (\(\uparrow\))} & \multirow{2}{*}{Latency (\(ms\downarrow\))} & \multirow{2}{*}{VRAM (MB\(\downarrow\))} \\
      \cline{2-6}
       & Rel. & Obj. & Pred. & Obj. & Pred. &  &  \\
      \hline
       FROSS \cite{hou2025fross} & 22.3 & 26.1 & 27.8 & 28.8 & 20.4 & 7 & - \\
       FROSS\(\dagger\) \cite{hou2025fross} & 22.3 & 25.3 & 27.5 & 27.6 & 12.6 & \cellcolor{best}\textbf{17} & \cellcolor{best}\textbf{1206} \\
       \hline
       Ours(\(n=128\)) & \cellcolor{secbest}32.4 & \cellcolor{secbest}28.0 & \cellcolor{secbest}34.6 & \cellcolor{secbest}29.6 & \cellcolor{secbest}16.5 & \cellcolor{secbest}22 & \cellcolor{secbest}1230 \\
       Ours(\(n=256\)) & \cellcolor{best}\textbf{36.9} & \cellcolor{best}\textbf{28.6} & \cellcolor{best}\textbf{39.5} & \cellcolor{best}\textbf{29.8} & \cellcolor{best}\textbf{18.6} & 23 & \cellcolor{secbest}1230 \\
    \bottomrule
    \end{tabular*}
    
    \label{tab:replicassg}
\end{table}

We present the main performance comparison for the top-1 recall and mean recall against other baseline methods in \cref{tab:main}. Our method surpasses all baselines in top-1 recall and object mRecall while maintaining competitive latency and comparable VRAM usage for real-time inference. More impressively, NoPA with 128 particles already surpasses all baselines. These results validate our hypothesis that the expressiveness of our continuous non-parametric formulation and our improved merging process contribute to the superior performance of NoPA. For ReplicaSSG in \cref{tab:replicassg}, NoPA surpasses FROSS across all metrics with a particularly significant margin for relationship recall (65.5\% increase).

\subsection{Qualitative Results}
\begin{figure}[t]
  \centering
  \includegraphics[width=\textwidth]{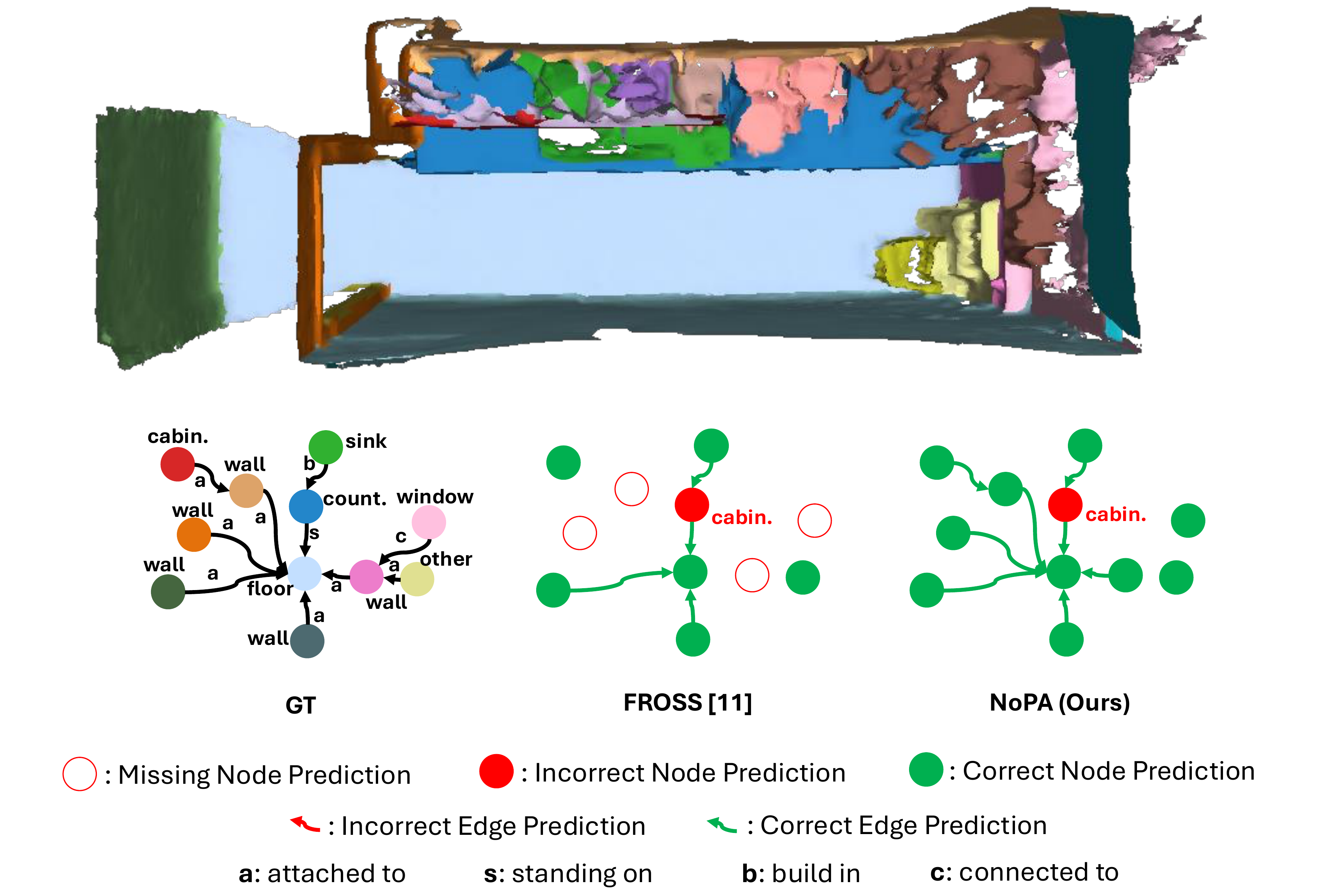}
  \caption{
  We compare the qualitative results between FROSS and our proposed approach for scene 321c867e from the 3DSSG dataset. The scene shows a kitchen from bird's eye view (BEV). FROSS fails to predict a majority of the wall background class. As a consequence, a majority of the predicate relationships are lost. Our method correctly classifies most objects, except for the counter instance, while correctly predicting the majority of predicate relationships.
  }
  
  \label{fig:qualitative2}
\end{figure}

We visualize the qualitative results compared to FROSS in \cref{fig:qualitative2}. Compared to FROSS, our approach can differentiate objects with thin structures such as the window class with a higher success rate. For ambiguous cases such as the misclassification of the counter instance, the failure is understandable since both methods rely on coarse approximations while differentiating between a counter and a cabinet is non-trivial.
Notably, NoPa correctly associates partial wall observations with the appropriate wall instance despite their textureless appearance. In contrast, this ambiguity poses a challenge for FROSS, which often produces incorrect merges and therefore exhibits poorer performance.
Additional qualitative results are provided in the supplementary material.

\subsection{Ablation Studies}
\begin{table}[t]
    \centering
    \caption{Ablation study on the test split of the 3DSSG dataset. NP. refers to the usage of our non-parametric particle set distribution instead of Gaussian distribution to represent objects in the scene. Merge. refers to the usage of our MMD-based merging approach. Prop. refers to the usage of our relationship propagation mechanism. The \textbf{Best} results are shown in bold.}
    
    \renewcommand{\arraystretch}{0.92}
    \setlength{\abovecaptionskip}{4pt}
    \setlength{\belowcaptionskip}{-4pt}
    
    \begin{tabular*}{\linewidth}{@{\extracolsep{\fill}}c|ccc|ccc|cc}
    \toprule
      \multirow{2}{*}{Method} & \multirow{2}{*}{NP.} & \multirow{2}{*}{Merge.} & \multirow{2}{*}{Prop.} 
      & \multicolumn{3}{c|}{Recall\% (\(\uparrow\))} 
      & \multicolumn{2}{c}{mRecall\% (\(\uparrow\))} \\
      \cline{5-9}
      & & & & Rel. & Obj. & Pred. & Obj. & Pred. \\
      \hline
       FROSS  & $\times$ & $\times$ & $\times$ & 25.7 & 60.6 & 30.7 & 62.4 & 17.7 \\
       \hline
       + NP. & $\checkmark$ & $\times$ & $\times$ & 17.6 & 66.1 & 20.8 & 64.8 & 11.1 \\
       + Merge. & $\checkmark$ & $\checkmark$ & $\times$ & 26.3 & \textbf{69.0} & 31.0 & \textbf{66.4} & 17.1 \\
       \hline
       Ours & $\checkmark$ & $\checkmark$ & $\checkmark$ & \textbf{53.2} & \textbf{69.0} & \textbf{61.4} & \textbf{66.4} & \textbf{29.4} \\
    \bottomrule
    \end{tabular*}
    
    \label{tab:ablation}
\end{table}

\begin{table}[t]
    \centering
    \caption{Comparison between different values of MMD threshold on the validation split of the 3DSSG dataset \(\delta_{\mathrm{MMD}}\). The \textbf{Best} results are shown in bold.}
    
    \renewcommand{\arraystretch}{0.92}
    \setlength{\abovecaptionskip}{4pt}
    \setlength{\belowcaptionskip}{-4pt}
    
    \begin{tabular*}{\linewidth}{@{\extracolsep{\fill}}c|ccccccc}
    \toprule
       \(\delta_{\mathrm{MMD}}\) & 0.6 & 0.65 & 0.7 & 0.75 & 0.8 & 0.85 & 0.9 \\
      \hline
       Rel.  & 36.3 & 47.9 & \textbf{53.7} & 51.3 & 46.4 & 46.5 & 46.3 \\
       Obj.  & \textbf{68.5} & 67.0 & 66.0 & 63.3 & 59.2 & 59.3 & 59.2 \\
       Pred. & 40.6 & 53.3 & \textbf{61.0} & 58.8 & 53.7 & 53.7 & 53.5 \\
    \bottomrule
    \end{tabular*}
    
    \label{tab:merge_comp}
\end{table}

To validate the contribution of each component of our proposed approach, we ablate each component in \cref{tab:ablation}. Replacing the Gaussian distribution with the particle set distribution increases recall for objects at the expense of predicate and relationship recall. The merging approach in FROSS for parametric representations clearly degrades performance when applied to our non-parametric representation due to fundamental incompatibilities between the two formulations. Integrating the distribution replacement with our tailored merging approach for non-parametric representations yields further improvements across all metrics.
The relationship propagation mechanism further boosts both relationship and predicate recall with no degradation in object recall since the position and spatial extent of all objects are unaffected by its use.

We analyze the effects of our merging approach in \cref{tab:merge_comp}. We vary the values of \(\delta_{\mathrm{MMD}}\) between the values of 0.6 and 0.9. We observe that a lower threshold improves object recall at the expense of predicate recall due to stricter merging criteria. Conversely, a higher threshold improves predicate and relationship recall with a trade-off in object recall. This corroborates the findings in \cite{hou2025fross}. The matching criteria discard predicted objects after merging that either lack sufficient overlap with the ground truth or excessively overlap with other matched objects due to their increased spatial extent. Relaxing the merging criterion produces more merged objects with a larger spatial extent, thereby reducing object recall.

Calculating MMD for particle set distribution similarity is superior to the approximation of the particle set distribution into a 3D Gaussian needed for the calculation of Hellinger distance especially for ambiguous cases. A theoretical explanation is the possible presence of misclassified objects from the predictions of the 2D SSG. The covariance calculated in Hellinger distance alone cannot capture the full distributional difference in object class that the use of MMD captures. This overreliance on covariance alone possibly leads to incorrect merges between objects from different classes which may explain the superiority of our approach. For more ablations and analysis, refer to the supplementary material.

\medskip
\noindent \textbf{Limitations.} \label{limit} Similar to prior works that lift 2D SSG to 3D, our approach heavily depends on the accuracy of the 2D SSG predictions from the pretrained models, especially for object prediction. The quality of 2D detections and relations cap the upper bound of performance for our approach.

\section{Conclusion}
We present \textbf{NoPA}, a non-parametric framework for online 3D scene graph generation from multi-view RGB-D observations. 
Our method replaces Gaussian object models with fixed-size particle sets that preserve geometric support while maintaining constant memory and runtime complexity. 
This design removes the restrictive ellipsoidal assumption and yields more stable multi-view object associations.
We introduce a distribution-level merging criterion based on MMD that compares particle sets directly in feature space and improves robustness under viewpoint variation and noisy predictions. 
A lightweight Hellinger distance pre-filter preserves efficiency by avoiding unnecessary MMD evaluations. 
We further propose a relationship propagation mechanism that recovers missing relations and improves global graph consistency.
Experiments show state-of-the-art performance on multiple online 3D scene graph benchmarks while maintaining competitive real-time efficiency. 
These results demonstrate that non-parametric object representations offer a practical alternative to parametric modeling for online 3D SSG generation.

\section*{Acknowledgments}
This research / project is supported by the National Research Foundation (NRF) Singapore, under its NRF-Investigatorship Programme (Award ID. NRF-NRFI09-0008), and the Tier 2 grant MOET2EP20124-0015 from the Singapore Ministry of Education.

\bibliographystyle{splncs04}
\bibliography{main}

\clearpage

\appendix 
\appendixtitle[NoPA: Non-Parametric Online 3D Scene Graph Generation]

In this supplementary material, we conduct more qualitative and quantitative analysis.
\begin{itemize}
    \item \ours is evaluated on additional quantitative and qualitative experiments in~\cref{sec:experiments} such as the per-class experiments (\cref{sec:per_class}), experiments using ground truth 2D SSG as input (\cref{sec:gt_ssg}), and additional qualitative results (\cref{sec:qualitative}).
    \item To justify our choice of representation type for objects, we analyze the effectiveness of \ours's non-parametric formulation compared to other parametric methods in~\cref{sec:parametric_compare}.
    \item To explain why relationship propagation works, we analyze the difference between the relationship propagation mechanism and an alternate merging based mechanism to recover relationships that are lost to under-merging in~\cref{sec:rel_propagation}.
    \item To understand our choice to focus on hard merge decisions, we analyze the impact of varying the margin band to determine how our definition of ambiguity affects \ours's performance in~\cref{sec:ambiguity}.
    \item To find the right balance between speed and performance, we analyze how the number of particles used to represent each object influences \ours's performance in~\cref{sec:num_particles}.
\end{itemize}

\section{Additional Experiments}
\label{sec:experiments}

\subsection{Per-class Experiments}
\label{sec:per_class}

We present the per-class experiments on the test split of the 3DSSG~\cite{wald2020learning} dataset in \cref{tab:perclass_obj_3dssg} and \cref{tab:perclass_rel_3dssg}. We reproduced FROSS~\cite{hou2025fross} and JointSSG~\cite{wu2023incremental} from their respective GitHub repositories. \ours excels across the different object categories with the best performance in all but three of the object classes. Specifically, \ours can better differentiate thin planar objects such as the \textit{picture} and \textit{window} classes compared to FROSS and JointSSG. For background classes such as \textit{floor} and \textit{wall}, FROSS' poorer performance can be attributed to subpar merging. This causes the resulting object to have low overlap with the ground truth. Conversely, our superior merging formulation allows us to correctly identify and merge these textureless categories into an object with corresponding large overlap with the ground truth to correctly tackle these challenging instances. 

For per-class predicate performance, both FROSS and \ours exhibits similar poor performance on rare classes inherited from their common RT-DETR-EGTR~\cite{Zhao_2024_CVPR, Im_2024_CVPR} backbone for 2D SSG prediction due to 3DSSG's severe class imbalance for predicate classes~\cite{wu2023incremental}. However, \ours excels in classifying relationships that co-occurs with objects that our superior merging formulation correctly identifies. \ours significantly outperforms all competing methods on the \textit{attached to} predicate class that is commonly found with the \textit{wall} object class. A similar phenomenon is shown between the \textit{standing on} predicate class with large furniture objects such as \textit{desk} and \textit{cabinet}. These observations can be verified visually in~\cref{sec:qualitative}. Although tackling the issue of class imbalance may further improve performance for \ours, we leave it to future work since solving class imbalance is not the main focus of this paper.

\begin{table*}[]
    \centering
    \caption{Comparison with state-of-the-art methods on the test split of the 3DSSG dataset for each object class. The \colorbox{best}{\textbf{Best}} results are highlighted.}
    \scalebox{0.58}{
    \begin{tabular}{c|cccccccccccccccccccc|c}
    \toprule
      Method   & bath. & bed & bkshf & cab. & chair & cntr. & curt. & desk & door & floor & ofurn & pic. & refri. & show. & sink & sofa & table & toil. & wall & wind. & Mean. \\
      \hline
       JointSSG~\cite{wu2023incremental} & \cellcolor{best} \textbf{100.0} & \cellcolor{best} \textbf{100.0} & 0.0 & 47.6 & \cellcolor{best} \textbf{67.4} & 25.0 & 57.6 & 20.0 & 50.0 & 90.0 & 10.7 & 14.3 & 0.0 & 0.0 & 42.9 & 55.0 & 54.1 & 80.0 & 68.0 & 25.0 & 45.4 \\
       FROSS~\cite{hou2025fross} & \cellcolor{best} \textbf{100.0} & 83.3 & \cellcolor{best} \textbf{28.6} & 53.0 & 61.3 & \cellcolor{best} \textbf{64.5} & \cellcolor{best} \textbf{77.4} & 25.0 & \cellcolor{best} \textbf{74.1} & 88.4 & 42.3 & 43.0 & \cellcolor{best} \textbf{50.0} & \cellcolor{best} \textbf{42.9} & \cellcolor{best} \textbf{76.7} & 75.0 & \cellcolor{best} \textbf{65.9} & \cellcolor{best} \textbf{100.0} & 57.0 & 40.2 & 62.4 \\
       \hline
       NoPA (Ours) & \cellcolor{best} \textbf{100.0}  & \cellcolor{best} \textbf{100.0} & \cellcolor{best} \textbf{28.6} & \cellcolor{best} \textbf{58.7} & 66.0 & \cellcolor{best} \textbf{64.5} & \cellcolor{best} \textbf{77.4} & \cellcolor{best} \textbf{37.5} & 69.0 & \cellcolor{best} \textbf{93.9} & \cellcolor{best} \textbf{44.4} & \cellcolor{best} \textbf{49.5} & \cellcolor{best} \textbf{50.0} & \cellcolor{best} \textbf{42.9} & 68.3 & \cellcolor{best} \textbf{76.3} & \cellcolor{best} \textbf{65.9} & \cellcolor{best} \textbf{100.0} & \cellcolor{best} \textbf{78.5} & \cellcolor{best} \textbf{57.5} & \cellcolor{best} \textbf{66.4} \\
    \bottomrule     
    \end{tabular}}
    \label{tab:perclass_obj_3dssg}
\end{table*}

\begin{table}[]
    \centering
    \caption{Comparison with state-of-the-art methods on the test split of the 3DSSG dataset for each predicate class. The \colorbox{best}{\textbf{Best}} results are highlighted.}
    \scalebox{0.85}{\begin{tabular}{c|ccccccc|c}
    \toprule
        Method & attached to & build in & connected to & hanging on & part of & standing on & supported by & Mean.\\
        \hline
        JointSSG~\cite{wu2023incremental} & 57.2 & 28.6 & \cellcolor{best} \textbf{29.4} & \cellcolor{best} \textbf{15.0} & \cellcolor{best} \textbf{20.0} & 0.4 & 11.4 & 23.1 \\
        FROSS~\cite{hou2025fross} & 26.0 & \cellcolor{best} \textbf{48.7} & 0.0 & 0.7 & 0.0 & 45.1 & 3.2 & 17.7 \\
        \hline
        NoPA (Ours) & \cellcolor{best} \textbf{73.2} & \cellcolor{best} \textbf{48.7} & 0.0 & 0.7 & 0.0 & \cellcolor{best} \textbf{71.4} & \cellcolor{best} \textbf{11.6} & \cellcolor{best} \textbf{29.4} \\
    \bottomrule
    \end{tabular}}
    
    \label{tab:perclass_rel_3dssg}
\end{table}

\begin{table*}[]
    \centering
    \caption{Comparison with state-of-the-art methods on the test split of the ReplicaSSG dataset for each object class. The \colorbox{best}{\textbf{Best}} results are highlighted.}
    \scalebox{0.68}{
    \begin{tabular}{c|ccccccccccccccccc|c}
    \toprule
      \multirow{2}{*}{Method}   & bag & bskt. & bed & bench & bike & book & botl. & bowl & box & cab. & chair & clock & cntr. & cup & curt. & desk & door & \multirow{2}{*}{Mean.} \\
       & lamp. & pil. & plant & plate & pot & rail. & scrn. & shlf. & shoe & sink & stand & table & toil. & towel & umb. & vase & wind. & \\
      \hline
       \multirow{2}{*}{FROSS} & \cellcolor{best} \textbf{25.0} & \cellcolor{best} \textbf{40.0} & 0.0 & 0.0 & 0.0 & \cellcolor{best} \textbf{2.2} & \cellcolor{best} \textbf{9.1} & 37.5 & 0.0 & \cellcolor{best} \textbf{4.8} & 71.0 & \cellcolor{best} \textbf{66.7} & 40.0 & \cellcolor{best} \textbf{28.6} & \cellcolor{best} \textbf{9.1} & 0.0 & 73.3 &  \multirow{2}{*}{ \textbf{27.6}} \\
       & \cellcolor{best} \textbf{16.7} & 43.4 & \cellcolor{best} \textbf{36.8} & \cellcolor{best} \textbf{31.2} & \cellcolor{best} \textbf{7.7} & 0.0 & 0.0 & 11.1 & \cellcolor{best} \textbf{8.3} & \cellcolor{best} \textbf{100.0} & 0.0 & \cellcolor{best} \textbf{66.7} & \cellcolor{best} \textbf{100.0} & 0.0 & \cellcolor{best} \textbf{66.7} & \cellcolor{best} \textbf{38.9} & \cellcolor{best} \textbf{3.3} & \\
       \hline
       \multirow{2}{*}{NoPA (Ours)} & \cellcolor{best} \textbf{25.0} & \cellcolor{best} \textbf{40.0} & 0.0 & 0.0 & 0.0 & 1.5 & \cellcolor{best} \textbf{9.1} & \cellcolor{best} \textbf{40.6} & \cellcolor{best} \textbf{8.0} & \cellcolor{best} \textbf{4.8} & \cellcolor{best} \textbf{87.0} & \cellcolor{best} \textbf{66.7} & \cellcolor{best} \textbf{60.0} & \cellcolor{best} \textbf{28.6} & \cellcolor{best} \textbf{9.1} & 0.0 & \cellcolor{best} \textbf{86.7} & \cellcolor{best} \\
       & \cellcolor{best} \textbf{16.7} & \cellcolor{best} \textbf{67.9} & \cellcolor{best} \textbf{36.8} & 25.0 & \cellcolor{best} \textbf{7.7} & 0.0 & 0.0 & \cellcolor{best} \textbf{22.2} & \cellcolor{best} \textbf{8.3} & \cellcolor{best} \textbf{100.0} & 0.0 & 55.6 & \cellcolor{best} \textbf{100.0} & 0.0 & \cellcolor{best} \textbf{66.7} & \cellcolor{best} \textbf{38.9} & 0.0 & \multirow{-2}{*}{\cellcolor{best} \textbf{29.8}}  \\
    \bottomrule     
    \end{tabular}}
    \label{tab:perclass_obj_replica}
\end{table*}

\begin{table}[]
    \centering
    \caption{Comparison with state-of-the-art methods on the test split of the ReplicaSSG dataset for each predicate class. The \colorbox{best}{\textbf{Best}} results are highlighted.}
    \scalebox{1.0}{\begin{tabular}{c|cccccccc|c}
    \toprule
        Method & above & against & attached to & in & near & on & under & with & Mean.\\
        \hline
        FROSS~\cite{hou2025fross} & 0.0 & 0.0 & 0.0 & 0.0 & 30.7 & 20.2 & 0.0 & 50.0 & 12.6 \\
        \hline
        NoPA (Ours) & 0.0 & 0.0 & 0.0 & 0.0 & \cellcolor{best} \textbf{45.1} & \cellcolor{best} \textbf{23.6} & 0.0 & \cellcolor{best} \textbf{80.0} & \cellcolor{best} \textbf{18.6} \\
    \bottomrule
    \end{tabular}}
    
    \label{tab:perclass_rel_replica}
\end{table}

\subsection{Ground Truth 2D SSG Experiments}
\label{sec:gt_ssg}
As implied in \cref{sec:per_class}, our approach heavily depends on the accuracy of the 2D SSG predictions from the pretrained models, especially for object prediction. 
Without accurate object classification in at least one frame, predicates that involve the misclassified object become incorrect or are entirely missed. This degrades performance for both the associated predicate classes and the affected object class.
A case in point is the poor performance of FROSS on the \textit{attached to} predicate class commonly found with the frequently misclassified \textit{wall} class. 

To investigate the degree of reliance on 2D SSG quality, we compare both FROSS and \ours with their oracle variants that take ground truth 2D SSG as input in~\cref{tab:gt_ssg}. With the ground truth 2D SSG, \ours experiences significant improvement across all metrics especially in predicate mRecall. This suggests that the performance of \ours scales with more accurate 2D SSG inputs while being more robust to inferior 2D SSG quality compared to FROSS.

\begin{table}[t]
    \centering
    \caption{Comparison on the usage of ground truth 2D SSG versus predicted 2D SSG on the test split of the 3DSSG dataset. + GT refers to the use of the oracle variant that takes in the ground truth 2D SSG as input.}

    \setlength{\abovecaptionskip}{4pt}
    \setlength{\belowcaptionskip}{-4pt}
    
    \scalebox{0.9}{\begin{tabular}{c|ccc|cc}
    \toprule
      \multirow{2}{*}{Model} & \multicolumn{3}{c|}{Recall\% (\(\uparrow\))} & \multicolumn{2}{c}{mRecall\% (\(\uparrow\))} \\
      \cline{2-6}
       & Rel & Obj. & Pred. & Obj. & Pred \\
      \hline
       FROSS~\cite{hou2025fross} & 25.7 & 60.6 & 30.7 & 62.4 & 17.7  \\
        FROSS~\cite{hou2025fross} + GT & \textbf{44.6 (+18.9)} & \textbf{77.2 (+16.6)} & \textbf{44.8 (+14.1)}& \textbf{83.7 (+21.3)}& \textbf{42.7 (+25.0)} \\
        \hline
        NoPA & 53.2 & 69.0 & 61.4 & 66.4 & 29.4  \\
        NoPA + GT & \textbf{84.4 (+15.4)} & \textbf{88.2 (+26.8)} & \textbf{85.0 (+18.6)}& \textbf{89.3 (+22.9)}& \textbf{80.8 (+51.4)}\\
    \bottomrule
    \end{tabular}}
    
    \label{tab:gt_ssg}
\end{table}

\subsection{More Qualitative Results}
\label{sec:qualitative}
\begin{figure}[t]
  \centering
  \includegraphics[width=\textwidth]{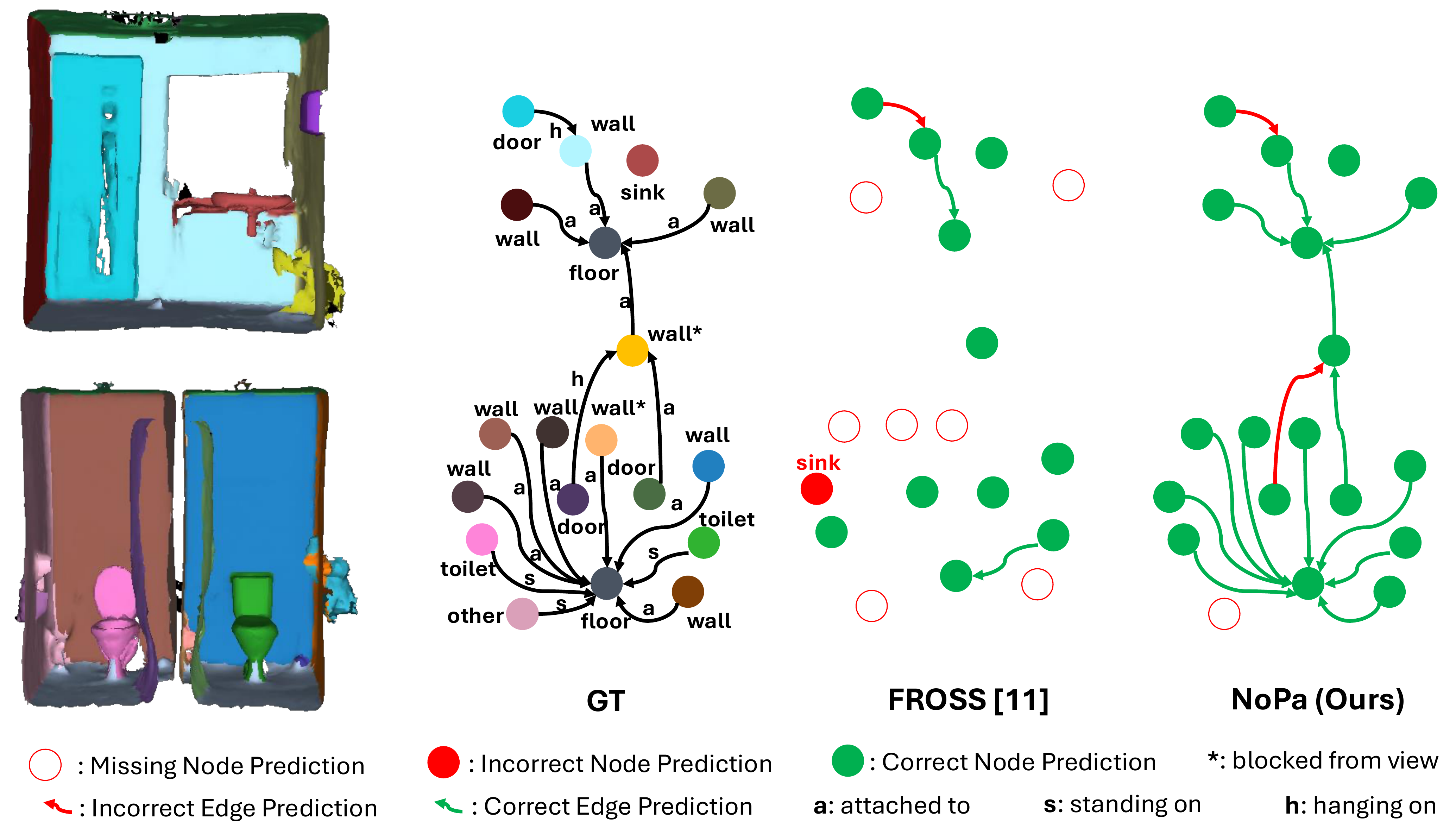}
  \caption{
  We compare the qualitative results between FROSS and our proposed approach for scene \texttt{ab835fae} from the 3DSSG dataset. The object instances denoted with a * are not visible from either viewpoint angles but are visible in the input images. FROSS fails to predict a majority of the \textit{wall} background class. Notably, FROSS has trouble differentiating the \textit{wall} class with the \textit{sink} class. FROSS also fails to predict a majority of the predicate relationships. Our method correctly classifies most objects while correctly predicting the majority of predicate relationships.}
  
  \label{fig:qualitative1}
\end{figure}

\begin{figure}[t]
  \centering
  \includegraphics[width=\textwidth]{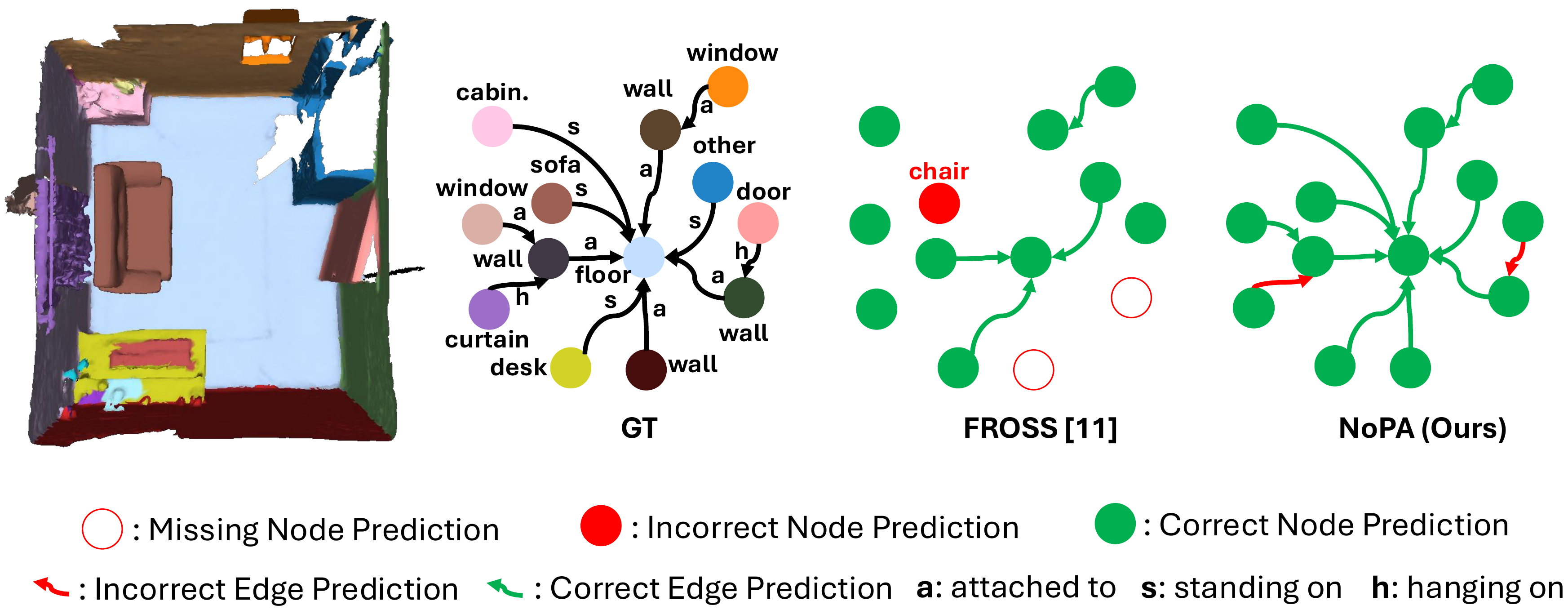}
  \caption{
  We compare the qualitative results between FROSS and our proposed approach for scene \texttt{c2d9933f} from the 3DSSG dataset. FROSS once again fails to predict a majority of the \textit{wall} background class. FROSS also misclassifies the \textit{sofa} instance as a \textit{chair} instance. Even though FROSS correctly classifies most objects, FROSS fails to predict the predicate relationships between most objects. Our method correctly classifies all objects while correctly predicting the majority of predicate relationships.}
  
  \label{fig:qualitative3}
\end{figure}

\begin{figure}[t]
  \centering
  \includegraphics[width=\textwidth]{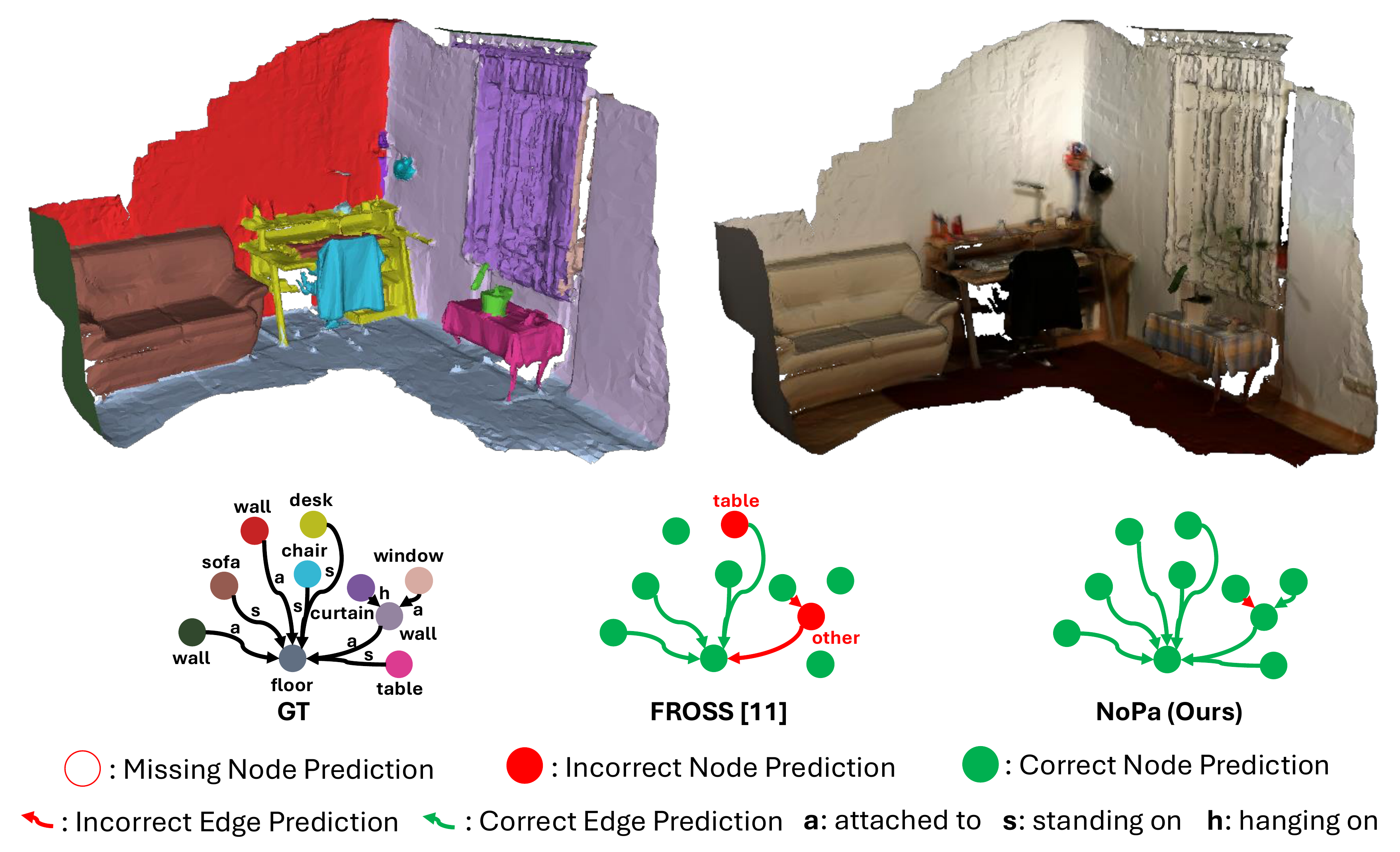}
  \caption{
  We compare the qualitative results between FROSS and our proposed approach for scene \texttt{5630cfe7} from the 3DSSG dataset. FROSS misclassifies the \textit{desk} object as a \textit{table} class. FROSS also misclassifies the \textit{wall} instance as an \textit{other furniture} instance. Because of the initial incorrect classification, all the relationships that are predicted with the \textit{wall} instance are misclassified or missing. Our method correctly classifies all objects while correctly predicting the majority of predicate relationships.}
  
  \label{fig:qualitative4}
\end{figure}

\cref{fig:qualitative1}, \cref{fig:qualitative3}, and \cref{fig:qualitative4} visualize more qualitative results on the 3DSSG dataset. As mentioned in~\cref{sec:per_class}, FROSS generally struggles to correctly identify the \textit{wall} class instances. 
FROSS still tends to omit relationships between the \textit{wall} instance and other object instances during classification, even when the \textit{wall} instance is correctly classified. \ours addresses this limitation through a more expressive representation that supports more reliable merging and preserves the predicted relationships.
Other classes with similar geometric shapes, such as \textit{sofa} and \textit{chair} or \textit{table} and \textit{desk}, are correctly classified by \ours while FROSS fails to classify accurately. In particular, both \ours and FROSS fail to predict the predicate \textit{hanging on} across many scenes. Since both \ours and FROSS rely on RT-DETR-EGTR for all possible predicate and object predictions, both methods are constrained by the initial 2D SSG predictions for each frame. 
With few correct predictions of the predicate \textit{hanging on} present in the initial 2D SSG predictions, both \ours and FROSS cannot manifest sufficient correct predicate predictions to overrule the majority for the final predicate prediction. Consequently, both methods fail to correctly classify the given predicate class. 
Nonetheless, \ours generalizes across diverse environments since each scene corresponds to a different room type. This result shows robustness to variations in object appearance and lighting conditions.

\section{Analysis}
\label{sec:analysis}

\subsection{Comparison with Other Representations}
\label{sec:parametric_compare}
Other than our non-parametric formulation, the objects can also be represented by other discrete representations such as the predicted 3D bounding boxes or point clouds lifted from the 2D bounding boxes. 

Reliable merging of 3D bounding boxes from sequential frame inputs requires careful design to maintain temporal consistency and avoid fragmented detections. Fusing the 3D bounding boxes from the candidate objects together requires a metric distance calculation between the two objects. Usually, this overlap is calculated via intersection over union (IoU). For our implementation, we leverage a calculation of IoU with a hard threshold of \(\delta_{IoU} = 0.1\). If the IoU between the local object candidate and the global object exceeds \(\delta_{IoU}\), we merge the two objects. Otherwise, we spawn the local object as a new global object. 

\smallskip
\noindent \textbf{Why it fails?} 
The main problem with using bounding boxes is the lack of any mechanism to reduce the size of the bounding box once expanded. 
This means that any outlier bounding box with an extended spatial extent can incorrectly influence the merged bounding box to expand to beyond the true object extent. Moreover, the background space is also captured in each bounding box. As a result, bounding boxes may be fused even when their overlap primarily corresponds to background regions rather than the underlying foreground objects. Given our strict criteria for ground truth matching, excessively large bounding boxes are filtered out. These large bounding boxes are generally composed of multiple merged objects with a larger number of accumulated relationships. 
As shown in~\cref{tab:representation}, filtering out such bounding boxes removes the associated relationships and reduces relationship recall.

For sparse point clouds obtained from the lifting of 2D bounding boxes, they cannot scale to real time performance if the object is present in a large number of frames in the scene since the number of points in each point cloud linearly increases. For our implementation, we employ a simple calculation of the point cloud overlap based on nearest neighbors with a similar hard threshold of \(\delta_{overlap} = 0.1\). If the overlap between the local object candidate and the global object exceeds \(\delta_{overlap}\), we merge the two objects. Otherwise, we spawn the local object as a new global object. 

\noindent \textbf{Why it fails?} Point clouds can capture the geometric detail of objects, but they are similarly affected by outliers. An outlier point results in the expansion of the point cloud to an incorrect extent. Most of the issues that plague 3D bounding boxes are also applicable to point clouds. The wrong fusion due to the presence of background space is mostly avoided since point clouds lifted from the bounding boxes largely only consist of the foreground object. 

In contrast to the prior two representations, our non-parametric representation is less affected by outliers since the resampling step after kernel density estimation tends to remove particles that are far away from the other particles. This ensures that the extent of the object is less likely to stretch beyond the ground truth object.

\begin{table}[t]
    \centering
    \caption{Comparison between different types of representations on the test split of the 3DSSG dataset. The \colorbox{best}{\textbf{Best}} results are highlighted.}
    
    \renewcommand{\arraystretch}{0.92} 
    \setlength{\abovecaptionskip}{4pt}
    \setlength{\belowcaptionskip}{-4pt}
    
    \begin{tabular*}{\linewidth}{@{\extracolsep{\fill}}c|ccc|cc|c|c}
    \toprule
      \multirow{2}{*}{Representation} & \multicolumn{3}{c|}{Recall\% (\(\uparrow\))} & \multicolumn{2}{c|}{mRecall\% (\(\uparrow\))} & \multirow{2}{*}{Latency (\(ms\downarrow\))} & \multirow{2}{*}{VRAM (MB\(\downarrow\))} \\
      \cline{2-6}
       & Rel & Obj. & Pred. & Obj. & Pred &  &  \\
      \hline
       Point Cloud & 23.3 & 65.0 & 27.4 & 64.2 & 16.2 & 71 & 1530 \\
       Bounding Box & 7.0 & 68.7 & 8.0 & 64.5 & 8.0 & \cellcolor{best} \textbf{26} & \cellcolor{best} \textbf{1204} \\
       \ours (Ours) & \cellcolor{best} \textbf{53.2} & \cellcolor{best} \textbf{69.0} & \cellcolor{best} \textbf{61.4} & \cellcolor{best} \textbf{66.4} & \cellcolor{best} \textbf{29.4} & 27 & 1206 \\
    \bottomrule
    \end{tabular*}
    
    \label{tab:representation}
\end{table}

\subsection{Analysis on Relationship Propagation}
\label{sec:rel_propagation}
The intuition behind relationship propagation is to recover relations that are missed throughout the incremental exploration of the scene. The caveat is that the relations must be present at some stage of the exploration process. This mechanism works under the assumption that neighboring object candidates with the same object class should have similar relations with similar objects. 
To group the object candidates into clusters, we reuse the precomputed MMD for merge decisions to avoid recomputation of an alternative metric over all particle sets to attain the affinity between object candidates. The affinity function is calculated as:
\vspace{-0.0em}
\begin{equation}
\mathbf{A} = [\max(0, 1 - \frac{d_{MMD}(\mathcal{X}_i(o), \mathcal{X}_j(o))}{2\delta_{MMD}})]_{i,j=1}^{n,n},
\end{equation}
which satisfies the criteria of boundedness and monotonicity thereby ensuring a well-formed formulation. If two segments possess low affinity score below affinity threshold \(\tau\), they are prevented from being grouped into the same cluster. This filtering mechanism avoids forming overly large clusters which improves amortized computational efficiency. 
The entire relationship propagation mechanism is described in \cref{algo:1}. 

\begin{algorithm}
    \caption{Relationship Propagation}
    \begin{algorithmic}[1]
        \State \(adj \in \mathbb{R}^{V \times V \times R} \text{ // R refers to the number of predicate classes}\)
        \State \(clusters \gets \{root: neigh(root)\}\) // V refers to the number of valid objects
        \For{$i = 1 \textbf{ to } C$} // C refers to the number of clusters
            \State \(cluster \gets clusters[i] \text{ // Obtained from Stage 2 decision rule}\)
            \For{\(j = 1 \textbf{ to } c_i\)} // \(c_i\) refers to number of objects in cluster i
                \For{\(k = 1 \textbf{ to } c_i-1\)}
                    \State // Accumulate relations for same cluster
                    \State \(r_j \gets adj[j]\)
                    \State \(r_k \gets adj[k]\)
                    \State \(r_c \gets r_j \cup r_k\)
                    \State \(adj[j] \gets r_c\)
                    \State \(adj[k] \gets r_c\)
                \EndFor
            \EndFor
        \EndFor
        \State \Return \(adj\)
    \end{algorithmic}
    \label{algo:1}
\end{algorithm}

\smallskip
\noindent \textbf{Why it works?} Neighboring objects from the same class tend to have similar relations with other objects. 
Similar to the concept of label propagation in semi-supervised learning, we are updating the labels based on information obtained from other data points. 
Our approach differs from label propagation in three ways: 1) We propagate relationships instead of object classes. 2) The nodes already contain prior information and are not unlabeled. 3) The propagation occurs in a single step rather than in multiple steps.

\smallskip
\noindent \textbf{Why not merge?} Merging also aggregates the relations from neighboring object candidates similar to relationship propagation. However, merging does not preserve the initial object candidates. As shown in~\cref{tab:propagation}, this preservation of object candidates heavily influences performance. If object candidates are merged into one instance when they belong to different instances, at least one matching with the ground truth is removed. Both matches may even be removed if the combined instance is filtered by our strict matching criteria. As a result, the relations that were previously augmented are still lost after merging. Consequently, the performance for all metrics degrades with the merging mechanism. This is also the reason why overly aggressive merging schemes fail, and \textbf{the correctness of each merge is critical to prevent the compounding of errors}.

\begin{wraptable}{r}{0.5\textwidth}
    \vspace{-0.8in}
    \centering
    \caption{Comparison between different types of relation aggregation methods on the test split of the 3DSSG dataset. The \colorbox{best}{\textbf{Best}} results are highlighted.}

    \scalebox{0.85}{\begin{tabular}{c|ccc|cc}
    \toprule
      \multirow{2}{*}{Mechanism} & \multicolumn{3}{c|}{Recall\% (\(\uparrow\))} & \multicolumn{2}{c}{mRecall\% (\(\uparrow\))} \\
      \cline{2-6}
       & Rel & Obj. & Pred. & Obj. & Pred \\
      \hline
       Merging & 25.2 & 67.8 & 29.4 & 66.3 & 15.5 \\
       Propagation & \cellcolor{best} \textbf{53.2} & \cellcolor{best} \textbf{69.0} & \cellcolor{best} \textbf{61.4} & \cellcolor{best} \textbf{66.4} & \cellcolor{best} \textbf{29.4} \\
    \bottomrule
    \end{tabular}}
    
    \label{tab:propagation}
    \vspace{-0.1in}
\end{wraptable}

\subsection{Analysis on Ambiguity}
\label{sec:ambiguity}
\begin{figure}[t]
  \centering
  \includegraphics[width=\textwidth]{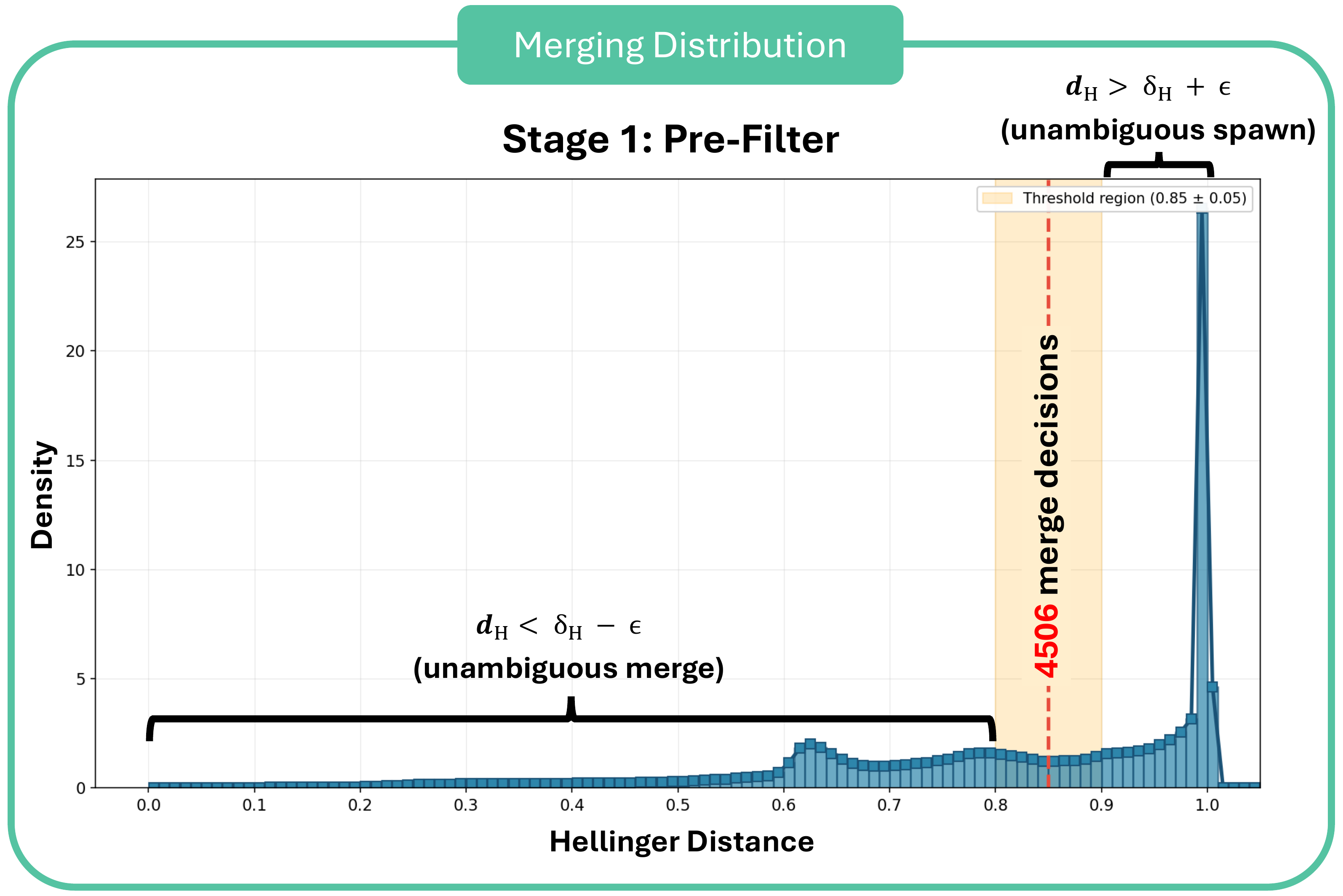}
  \caption{
  We show the distribution of merge decisions in the test split of the ReplicaSSG dataset according the Hellinger distance calculated from fitting the Gaussian distribution on \ours's particles. Even in the narrow margin band between \(\delta_H - \epsilon \leq d_H \leq \delta_H + \epsilon\) where \(\delta_H = 0.85\) and \(\epsilon = 0.05\), there exists a substantial number of merge decisions that requires sensitive MMD calculation beyond moments matching.}
  
  \label{fig:hellinger}
\end{figure}

\cref{fig:hellinger} shows the distribution of merge decisions that are considered ambiguous in the ReplicaSSG dataset. 
Even a small number of incorrect decisions within this margin band can cascade and lead to the compounding of errors that degrades \ours's performance. 
\cref{tab:ambiguity} presents the performance of \ours with different margin bands that correspond to varying levels of ambiguity. \(\epsilon = 0\) signifies the exclusive use of the Hellinger's distance for all fitted Gaussians on the particle sets of object candidates without additional MMD support. \ours attains the best performance with \(\epsilon = 0.05\). 

\smallskip
\noindent \textbf{Remarks.} Although MMD might be the superior choice when used for ambiguous merges, it loses its effectiveness when used for merging decisions that are clear cut according to the Hellinger distance. MMD compares the full distributional distance between two object distributions in a reproducing kernel Hilbert space. However, it does not explicitly encode the Euclidean distance between objects that the Hellinger distance encodes. The larger the margin band, the more likely that MMD may mistakenly merge two object candidates with a large Euclidean distance. Focusing on the narrow margin band allows MMD to provide more reliable merge decisions compared to relying on pure moment matching.

\begin{table}[t]
    \centering
    \caption{Comparison between different values of \(\epsilon\) on the test split of the ReplicaSSG dataset. The \colorbox{best}{\textbf{Best}} and \colorbox{secbest}{Second Best} results are highlighted, respectively.}

    \begin{tabular}{c|ccc|cc|c}
    \toprule
      \multirow{2}{*}{\(\epsilon\)} & \multicolumn{3}{c|}{Recall\% (\(\uparrow\))} & \multicolumn{2}{c|}{mRecall\% (\(\uparrow\))} & \multirow{2}{*}{Latency (\(ms\downarrow\))} \\
      \cline{2-6}
       & Rel & Obj. & Pred. & Obj. & Pred &   \\
      \hline
       0 & 16.2 & 27.4 & 17.2 & 29.0 & 7.9 &  \cellcolor{best} \textbf{18} \\
       0.05 & \cellcolor{best} \textbf{36.9} & \cellcolor{best} \textbf{28.6} & \cellcolor{best} \textbf{39.5} & \cellcolor{best} \textbf{29.8} & \cellcolor{best} \textbf{18.6} & \cellcolor{secbest}23 \\
       0.1 & \cellcolor{secbest}31.7 & \cellcolor{secbest}27.5 & \cellcolor{secbest}34.0 & \cellcolor{secbest}29.2 & \cellcolor{secbest}15.9 &  24 \\
       0.15 & 31.1 & 26.0 & 33.3 & 28.5 & 15.7 &  26 \\
    \bottomrule
    \end{tabular}
    
    \label{tab:ambiguity}
\end{table}

\subsection{Analysis on Number of Particles}
\label{sec:num_particles}
\begin{table}[t]
    \centering
    \caption{Comparison between different number of particles in each particle set on the validation split of the 3DSSG dataset. The \colorbox{best}{\textbf{Best}} and \colorbox{secbest}{Second Best} results are highlighted, respectively.}

    \begin{tabular}{c|ccc|cc|c|c}
    \toprule
      \multirow{2}{*}{\(n\)} & \multicolumn{3}{c|}{Recall\% (\(\uparrow\))} & \multicolumn{2}{c|}{mRecall\% (\(\uparrow\))} & \multirow{2}{*}{Latency (\(ms\downarrow\))} & \multirow{2}{*}{VRAM (MB\(\downarrow\))} \\
      \cline{2-6}
       & Rel & Obj. & Pred. & Obj. & Pred &  &  \\
      \hline
       64 & 50.8 & 63.3 & 57.5 & 52.8 & 26.8 & \cellcolor{best} \textbf{26} & \cellcolor{best} \textbf{1206} \\
       128 & 51.3 & \cellcolor{best} \textbf{66.7} & 58.8 & \cellcolor{secbest}56.1 & 27.8 & \cellcolor{secbest}27 & \cellcolor{best} \textbf{1206} \\
       256 & \cellcolor{best} \textbf{53.7} & \cellcolor{secbest}66.0 & \cellcolor{best} \textbf{61.0} & \cellcolor{best} \textbf{56.6} & \cellcolor{secbest}28.3 & 29 & \cellcolor{best} \textbf{1206} \\
       512 & \cellcolor{secbest} 53.0 & 65.3 & \cellcolor{secbest}60.3 & 55.2 & \cellcolor{best} \textbf{29.4} & 30 & \cellcolor{secbest} 1207 \\
    \bottomrule
    \end{tabular}
    
    \label{tab:particle_size}
\end{table}

Since the theoretical runtime of the Stage 2 decision rule using MMD is proportional to the square of the number of particles \(O(n) \propto n^2\), we expect the runtime to scale quadratically with the number of particles. However, since we only apply MMD to ambiguous merge decisions which only covers a smaller subset of merge decisions, the runtime cost is amortized. Empirically, as seen in~\cref{tab:particle_size}, we found that  
the runtime of \ours does not increase
substantially as the number of particles increases. \(n=256\) strikes the right balance in the tradeoff between speed and performance.

\smallskip
\noindent \textbf{Remarks.} Counterintuitively, a larger number of particles does not always lead to an increase in performance. \ours with \(n=512\) particles performs worse than \(n=256\). One possibility might be that the effective number of particles required to represent a single object is less than 512. The additional particles above \(n=256\) may be replicating the same information as the prior particles without adding extra context. 
Adding more particles may even be counterproductive as it can produce noisier object representations and reduce performance.

\end{document}